\documentclass[pdflatex,sn-basic]{sn-jnl}

\jyear{2024}\theoremstyle{thmstyleone}

\theoremstyle{thmstyletwo}

\theoremstyle{thmstylethree}

\usepackage{tabularx}
\newcolumntype{L}[1]{>{\raggedright\let\newline\\\arraybackslash\hspace{0pt}}p{#1}}
\newcolumntype{C}[1]{>{\centering\let\newline\\\arraybackslash\hspace{0pt}}p{#1}}
\newcolumntype{R}[1]{>{\raggedleft\let\newline\\\arraybackslash\hspace{0pt}}b{#1}}

\usepackage{epstopdf}

\begin{document}

\title[KD \& IG]{Model compression using knowledge distillation with integrated gradients}

\author[1]{\fnm{David E.} \sur{Hernandez}, \dgr{B.S.}}\email{david.hernandez@nordlinglab.org (0009-0004-4392-6927)}

\author[1]{\pfx{Dr} \fnm{Jose Ramon} \sur{Chang}, \dgr{MSc, PhD}}\email{jose.chang@nordlinglab.org (0000-0001-5587-7828)}

\author*[1]{\pfx{Dr} \fnm{Torbj{\"o}rn E. M.} \sur{Nordling}, \dgr{MSc, PhD}}\email{torbj{\"o}rn.nordling@nordlinglab.org (0000-0003-4867-6707)}

\affil[1]{\orgdiv{Department of Mechanical Engineering}, \orgname{National Cheng Kung University}, \orgaddress{\street{No. 1 University Rd.}, \city{Tainan} \postcode{701}, \country{Taiwan}}}

\abstract{
Model compression is critical for deploying deep learning models on resource-constrained devices. We introduce a novel method enhancing knowledge distillation with integrated gradients (IG) as a data augmentation strategy. 
Our approach overlays IG maps onto input images during training, providing student models with deeper insights into teacher models' decision-making processes.
Extensive evaluation on CIFAR-10 demonstrates that our IG-augmented knowledge distillation achieves 92.6\% testing accuracy with a 4.1x compression factor—a significant 1.1 percentage point improvement (p $<$ 0.001) over non-distilled models (91.5\%). 
This compression reduces inference time from 140ms to 13ms. Our method precomputes IG maps before training, transforming substantial runtime costs into a one-time preprocessing step.
Our comprehensive experiments include: (1) comparisons with attention transfer, revealing complementary benefits when combined with our approach; (2) Monte Carlo simulations confirming statistical robustness; (3) systematic evaluation of compression factor versus accuracy trade-offs across a wide range (2.2x-1122x); and (4) validation on an ImageNet subset aligned with CIFAR-10 classes, demonstrating generalisability beyond the initial dataset.
These extensive ablation studies confirm that IG-based knowledge distillation consistently outperforms conventional approaches across varied architectures and compression ratios. 
Our results establish this framework as a viable compression technique for real-world deployment on edge devices while maintaining competitive accuracy.
}


\maketitle

\section{Introduction}\label{sec:introduction}\label{sec:intro}

Deploying Deep Learning (DL) models in real-time applications is challenging due to their high computational demands, particularly on edge devices such as smartphones and IoT systems \citep{Szegedy2017Inceptionv4IA, deng2020model, krishnamoorthi2018quantizing}.
Traditional DL models, while effective, often exceed the resource capacities of these devices, causing issues like increased latency, higher energy consumption, larger memory requirements, and privacy risks \citep{han2015deep, howard2017mobilenets, sze2017efficient}.

LLMs like GPT-4 and DeepSeek-R1 achieve remarkable performance but are computationally expensive, making efficient model compression essential for resource-limited environments \citep{chen2024smallmodels}.
This highlights a critical research gap: while significant attention has focused on developing large models, the systematic improvement of compression techniques for practical deployment remains under-explored.
Our work addresses this gap by introducing an enhanced Knowledge Distillation (KD) framework that maintains model interpretability while achieving significant compression.

The paper is structured as follows: Section \ref{sec:methodology} details the methodology, including the teacher-student framework, KD approach, Attention Transfer (AT), and the incorporation of integrated gradients for data augmentation.
Section \ref{sec:experiments} explains the experimental procedure, including the dataset used and the hyperparameter search for KD and AT. 
Section \ref{sec:results} presents the experimental results, demonstrating the efficacy of our technique using MobileNet-V2 on the CIFAR-10 dataset. 
Finally, Section \ref{sec:conclusions} concludes by discussing potential future research directions in model compression for edge AI applications and the broader implications of our findings.

\subsection{Model Compression}\label{sec:model_compression}
Model compression techniques are generally divided into four categories: model pruning, parameter quantisation, low-rank factorisation, and knowledge distillation.
Table~\ref{tab:compression_techniques} summarises these methods, which address various needs such as reducing model size, improving computational speed, or maintaining accuracy under constraints \citep{cheng2017survey, molchanov2016pruning, courbariaux2016binarized, jaderberg2014speeding}. 

Recent reviews \citep{wang2024model, wang2020apq, liu2022bringing} highlight that while methods like pruning and quantisation often require a trade-off between accuracy and efficiency, KD offers a more balanced solution.
By transferring knowledge from a larger `teacher' model to a smaller `student' model, KD enables the student to retain much of the performance of the teacher while significantly reducing computational requirements.
This makes KD particularly effective for maintaining both efficiency and performance in resource-constrained environments \citep{Hinton2015}.

\begin{table*}
\begin{center}
\caption{Most widely-used model compression techniques.}
  \label{tab:compression_techniques}
\begin{tabular}{m{2cm} m{4cm} m{5cm}}
\hline
Compression Technique & Mechanism & Application \\ \hline
Model Pruning & Elimination of redundant components in pre-trained networks. & Modest computational speedups, 2-3 times \citep{wang2020pruning, xia2022structured}. \\
Parameter Quantisation & Conversion of floating-point values to discrete integers. & Suitable for hardware accelerators like Tensor Core from NVIDIA \citep{krishnamoorthi2018quantizing, markidis2018nvidia, pan2018multilevel}. \\
Low-Rank Factorisation & Reduction of matrix ranks via techniques like SVD. & Minimises memory and computational demands \citep{idelbayev2020low}. \\
Knowledge Distillation & A complex ``teacher" network trains a simpler ``student" network with soft outputs. & Enhances the effectiveness of the student model beyond hard labels \citep{Hinton2015}. \\\hline
\end{tabular}
\end{center}
\end{table*}

\subsection{Explainable AI} \label{sec:XAI}
Explainable AI (XAI) techniques seek to make machine learning models transparent and interpretable for human users.
While various XAI approaches exist, such as LIME \citep{ribeiro2016lime} and SHAP \citep{lundberg2017shap}, we focus on Integrated Gradients (IG) and attention mechanisms for their particular advantages in knowledge distillation.
These techniques not only provide insights into model decision-making but also offer practical benefits for guiding compression.

Interpretability is especially crucial in model compression, as practitioners need to verify that compressed models maintain both accuracy and fidelity to the original decision-making process.
This is particularly important in high-stakes domains such as healthcare, where compressed models deployed on edge devices must remain explainable.
For example, a compressed model for pneumonia detection from chest X-rays should highlight the same suspicious regions as the full model to assist clinicians in validating AI recommendations.

In our approach, we combine these interpretability tools with knowledge distillation. Attention mechanisms ensure the student model learns to focus on the same regions as the teacher, while IG provides pixel-level attribution maps that guide feature learning.
This integration of XAI with knowledge distillation provides both performance benefits and improved interpretability, which we quantify in our experimental results.

\subsection{Previous Works} \label{sec:lit_review}
\begin{figure}[tb]
  \centering
  \includegraphics[width=\linewidth]{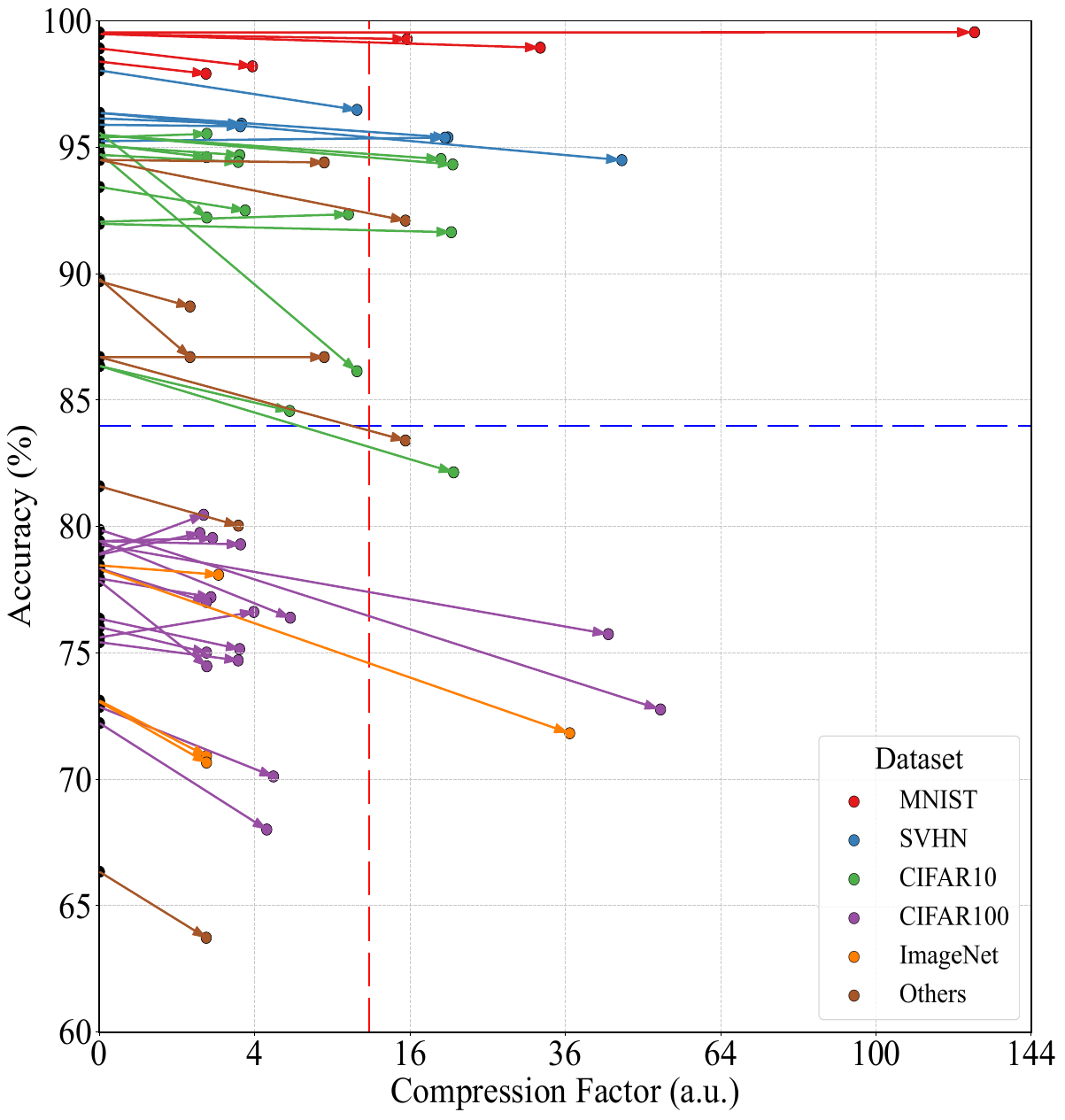}
  \caption{Difference in accuracy between the teacher model and the student model, as a function of the compression factor for the studies included in our review.
  The black dots are the teacher accuracies linking to the performance of the student models.
  A longer and flatter line means better results.
  Articles that did not report a compression factor were excluded. 
  The red dashed line represents the mean compression factor of 11.41.
  The blue dashed line represents the mean accuracy of 83.46. 
  Each marker colour represents a different dataset.
   }
  \label{fig:plot_studies}
\end{figure}

Our literature analysis examined model compression through knowledge distillation in image classification from 2017 onwards, focusing on compression factors and accuracy metrics across multiple datasets. 
The studies revealed a complex relationship between model compression and performance, with compression factors ranging from 1.4x to 127x and varied accuracy impacts ranging from -8.63\% to +1.54\%.
Figure~\ref{fig:plot_studies} shows the relationship between compression factors and accuracy changes across different studies and datasets. 
Clear trade-offs between model size reduction and performance emerge, with studies like \citet{chen2019data} showing that outcomes vary significantly with dataset complexity. 
Our analysis found a weak negative correlation $(r = -0.114, p = 0.414)$ between compression factor and accuracy loss, indicating that higher compression doesn't consistently lead to larger accuracy drops—some highly compressed models maintain strong performance while others experience significant degradation.
Table~\ref{tab:studies_statistics} presents statistical data across different datasets, revealing that simpler datasets achieve higher compression with minimal accuracy loss (MNIST: 44.7x average compression, -0.37\% median accuracy change), while complex datasets show more modest compression (ImageNet: 7.5x average, -1.40\% median accuracy change). 
Notably, some studies report student models outperforming their teachers in specific tasks \citep{ashok2017n2n, gou2023hierarchical}, suggesting that distillation can sometimes refine model performance beyond simple compression or that the teacher was not properly trained.
Recent developments have focused on more sophisticated distillation techniques to better balance these trade-offs. 
Studies such as \citet{gou2022multilevel, gou2023hierarchical} introduced multi-level and hierarchical distillation, offering finer control over compression-accuracy balance. 
\citet{choi2020data} explored adaptive distillation strategies that dynamically adjust based on task complexity, improving performance on challenging datasets.
Most relevant to our work, \citet{wu2023ad} utilised Integrated Gradients to transfer attribution-based knowledge in NLP tasks. 
While their approach incorporated IG in the loss function for BERT models, our research applies a different methodology to image classification using MobileNet-V2. 
We adapt IG as a data augmentation technique rather than an explicit loss term, guiding the student model toward critical focus areas within images. 
By precomputing IG for the dataset, we significantly reduce computational demands during training. 
Our methodology also evaluates knowledge distillation across student models of varying compression rates, providing insights into scalability and adaptability.

In this work we make the following contributions:
\begin{itemize}
    \item Propose a novel model compression method using integrated gradients to guide the learning of the smaller model and compare it to attention transfer.
    \item Introduce a more standardised approach to evaluate the performance of model compression algorithms.
    \item Perform Monte Carlo simulations showing that the improvements by KD, AT, and IG are statistically significant even under variation of the training data.
\end{itemize}

\begin{table*}[tb]
\begin{center}
\caption{Mean and median for the change in accuracy from teacher to student ($\Delta$Accuracy) and Compression Factor (CF) of different datasets.} \label{tab:studies_statistics}
  \begin{tabular}{lccccc}
\hline
  \multirow{2}{*}{Dataset} & \multicolumn{2}{c}{$\Delta$Accuracy} & \multicolumn{2}{c}{CF} & No. of\\ 
& Mean & Median & Mean & Median  & Articles\\\hline
CIFAR-10 & -0.98 & -0.61 & 8.88 & 4.76 & 14\\
CIFAR-100 & -1.73 & -1.27 & 9.03 & 2.62 & 14\\
SVHN & -0.75 & -0.59 & 17.14 & 15.40 & 6\\
ImageNet &-2.07& -1.40& 7.53& 2.12& 6\\
MNIST & -0.36 & -0.37 & 44.71 & 23.96 & 4\\
MARKET-1501 &-1.87&-2.40&8.42&8.38& 3 \\
Tiny-ImageNet &-2.76&-2.76&2.59&2.59 & 2\\
DukeMTMCreID &-1.65&-1.65&11.96&11.96 & 2\\
CelebA &-1.56& -1.56&3.20& 3.20& 1\\
CUHK03 &-1.00&-1.00&1.36&1.36 & 1\\
Caltech-256 &-2.94&-2.94&3.12&3.12 & 1\\\hline
\end{tabular}
\end{center}
\end{table*}

\section{Methodology} \label{sec:methodology}
\begin{figure}
  \centering
  \includegraphics[width=\linewidth]{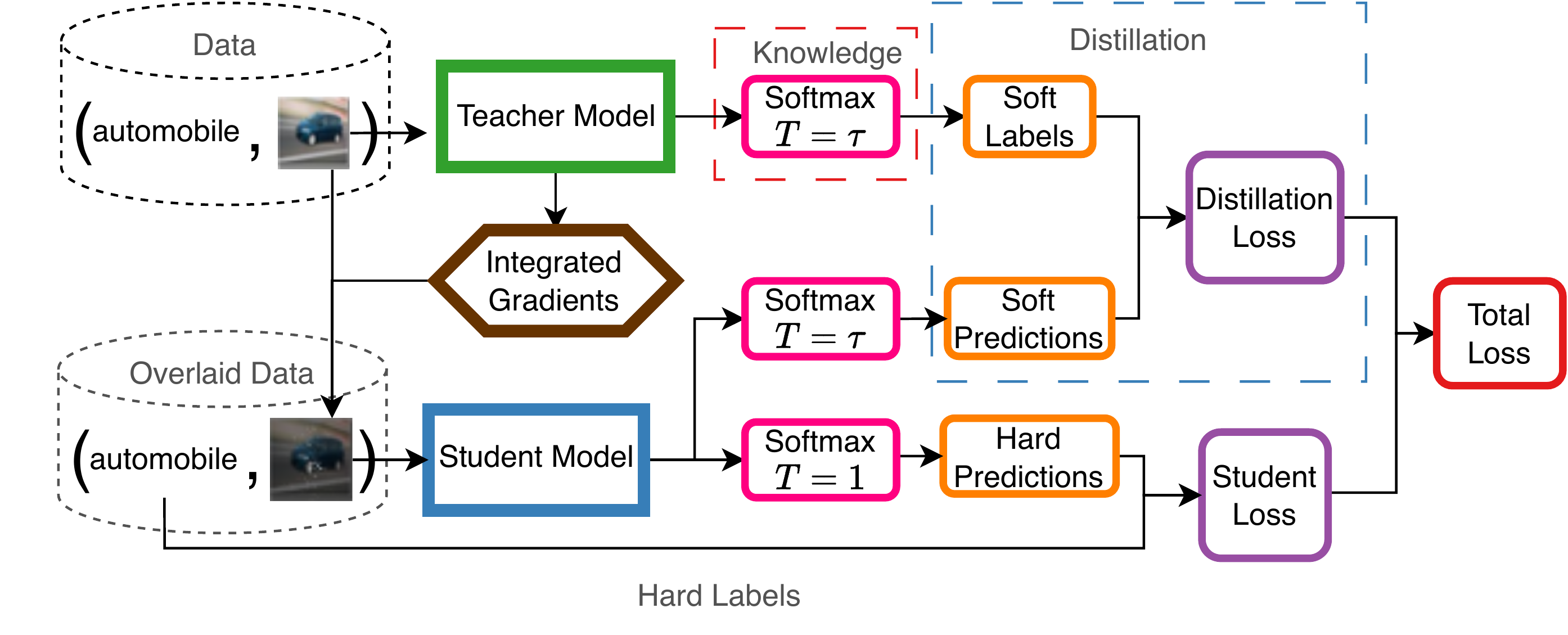}
  \caption{Knowledge distillation process using integrated gradients for data augmentation. 
  The teacher model (green) employs a temperature hyperparameter $T=\tau$ where $\tau > 1$ in its softmax function to produce soft targets, which, along with the hard labels from the dataset, guide the training of the student model (blue). 
  Integrated gradients (brown) are overlaid with the original images to generate enhanced data that focuses critical features that the student model should use during training.
}
  \label{fig:kd_ig_flow}
\end{figure}

Our approach combines three key components to achieve efficient model compression while maintaining interpretability: KD for transferring model knowledge, AT for preserving spatial understanding, and IG for feature-level guidance.
Figure~\ref{fig:kd_ig_flow} provides an overview of how these components work together in our framework.
We first describe each component individually, then detail their integration and implementation.

\subsection{Knowledge Distillation Framework}
Knowledge Distillation transfers information from a teacher model to a student model through two primary mechanisms: (1) soft targets generated by the teacher model with temperature scaling and (2) intermediate representations that capture the internal processing of the teacher.
The core mechanism leverages `softened' output probabilities that reflect the confidence levels of the teacher model across all classes, revealing the relational structure between categories.
For example, when classifying an image as `car', the teacher might assign meaningful probabilities to visually similar classes like `truck', helping the student understand both certainty and ambiguity in predictions.

The temperature hyperparameter in the softmax function controls this softening process, with higher values broadening the output distribution to prevent overconfidence and reveal more of the uncertainty from the teacher.
The objective function for standard knowledge distillation is defined as:

\begin{align} \label{eq:weighted_loss}
L_{KD} = (1 - \alpha) L_\mathcal{H}+ \alpha L_\mathcal{KL}.
\end{align}

\noindent Here $L_\mathcal{H}$ represents the cross-entropy loss between the student predictions and the ground-truth hard labels, and $L_\mathcal{KL}$ is the Kullback-Leibler divergence between the softened output distributions of the teacher and student, i.e. soft labels, scaled by temperature $T \in [1,20]$. 
The hyperparameter $\alpha$ weights the two signals. 
For details, see~\citet{gou2021knowledge}.

We selected MobileNetV2 as our teacher model due to its optimal balance between accuracy (93.9\% on CIFAR-10) and efficiency (2.2M parameters).
Its inverted residual architecture provides a feature-rich structure ideal for knowledge transfer while maintaining practical deployment potential in resource-constrained environments.

\subsection{Attention Transfer}
Building upon the base KD framework, we incorporate attention transfer to ensure the student model learns to focus on the same important regions as the teacher.
In neural networks, attention mechanisms reveal which regions of the input the network prioritises during decision-making.
These attention maps provide insights into the reasoning process of the model and can be derived from activations at various layers within the network.
When integrating attention transfer, we extend the objective function to include an attention loss term that aligns the spatial focus of the student with that of the teacher:

\begin{align}
L_{Total} = (1 - \alpha) L_\mathcal{H}+ \alpha L_\mathcal{KL} + \gamma L_{\text{AT}} \label{eqn:totalloss}
\end{align}

\noindent where $L_{\text{AT}} = \| A_S - A_T \|^2_2$ represents the Mean Squared Error between the $L_2$-normalized student and teacher attention maps, derived from the middle layer activations, and $\gamma$ controls the weighting of this attention alignment in the overall loss function.
A few representative attention maps are visualised in Figure S1 (Online Resource 1).

\subsection{Integrated Gradients as Data Augmentation}
\begin{figure}
  \centering
  \includegraphics[width=\linewidth]{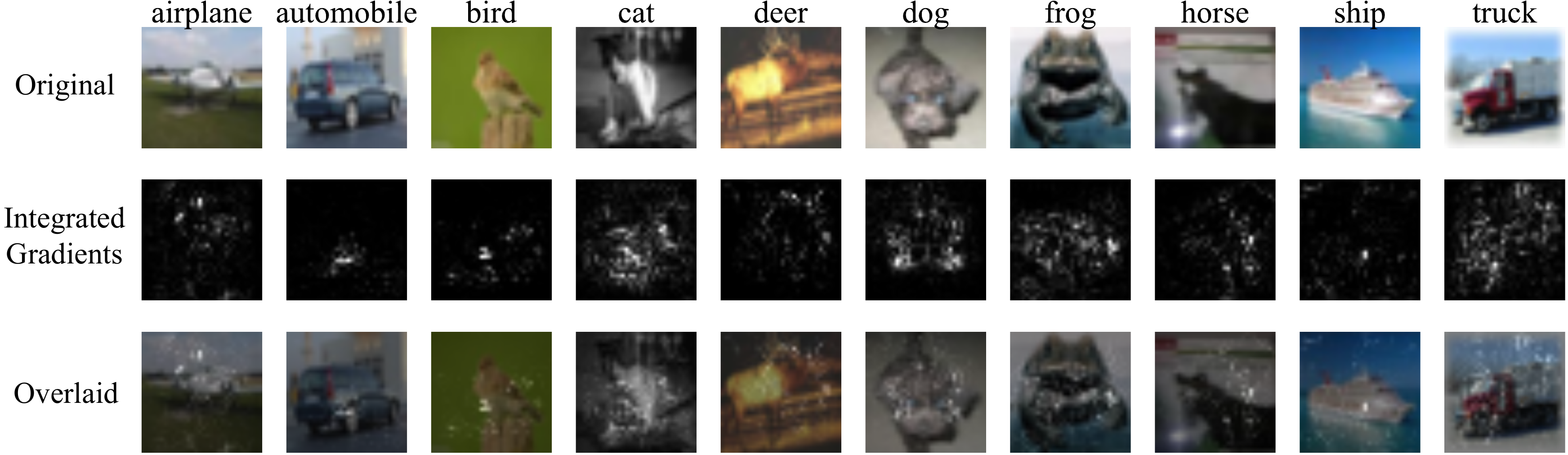}
  \caption{Implementation of IG as a data augmentation technique on CIFAR-10. 
  The top row shows the original images from various classes. 
  The middle row displays the Integrated Gradients, highlighting areas of the image that significantly influence the predictions of the teacher model. 
  The bottom row presents the overlaid images, which combine the original images with their respective integrated gradients to emphasise regions of interest, so that the student can more easily focus on these influential areas.
}
  \label{fig:augment_class}
\end{figure}

To further enhance the knowledge transfer process, we introduce IG-based data augmentation, as defined in~\citet{sundararajan2017axiomatic}.
Integrated gradients provide pixel-level attribution maps that identify which input features most influence the decisions of the model.
The IG for an input feature $i$ of the image is calculated as:

\begin{align}
IG_i(x) = (x_i - x'_i) \int_{\beta=0}^{1} \frac{\partial F(x' + \beta(x-x'))}{\partial x_i}d\beta
\end{align}

\noindent where $x \in [0, 1]^{C \times H \times W}$ is the input image tensor of shape channels (C), height (H), and width (W), $x'$ the baseline (typically a zero tensor of identical shape), $F$ the model function, and $\beta$ a scaling hyperparameter that transitions from baseline to input.
Rather than using IG as an explicit loss term, we implement it as a data augmentation technique, where IG maps are overlaid onto the original images with a controlled probability, as illustrated in Figure~\ref{fig:augment_class}.
The IG augmentation is done in three steps (scaling, normalisation to $[0,1]$, image overlay):

\begin{align}
IG_{\text{scaled}}(x) &= IG(x)^s \quad \text{with scale factor}\;s \sim \exp(\mathcal{U}[\ln(1), \ln(2)]),\\
\hat{IG}(x) &= \frac{IG_{\text{scaled}}(x) - \min(IG_{\text{scaled}}(x))}{\max(IG_{\text{scaled}}(x)) - \min(IG_{\text{scaled}}(x))}, \\
x_{\text{agumented}} &= 
\begin{cases} 
0.5 \cdot x + 0.5 \cdot \hat{IG} & \text{with probability}\; p, \\ 
x & \text{otherwise}.
\end{cases}
\end{align}

\begin{figure}[ht] 
  \centering 
  \includegraphics[width=\textwidth]{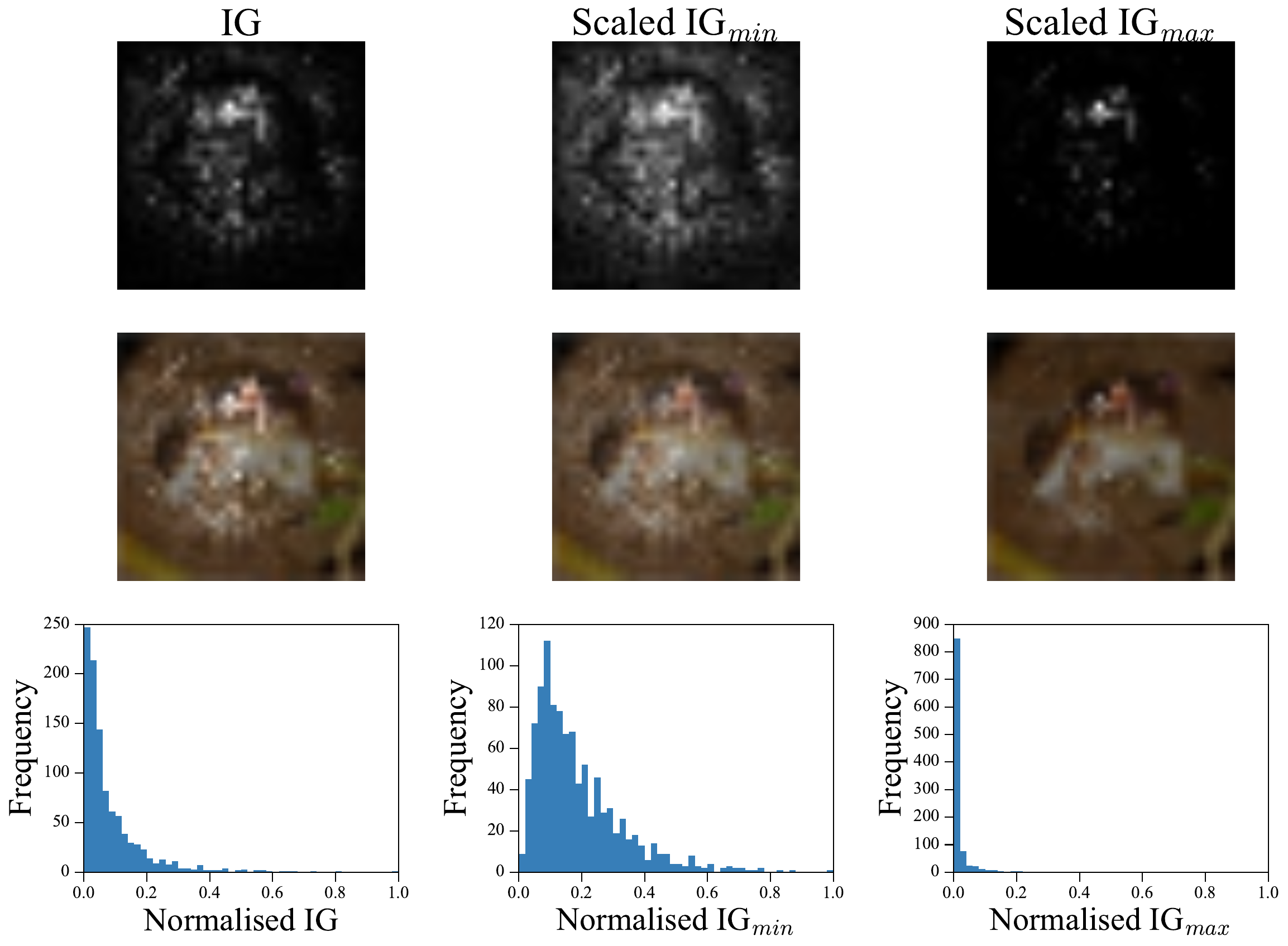} 
  \caption{Distribution of normalised IG using different scaling techniques.
\textit{Top row:} Grayscale attribution maps for standard normalised IG (left), IG with minimum log scale factor 0.6 (I$\text{G}_{min}$, middle), and IG with maximum log scale factor 2.0 (I$\text{G}_{max}$, right).
\textit{Middle row:} The same attribution maps overlaid on the original image, showing how different scaling affects feature visibility.
\textit{Bottom row:} Frequency histograms of attribution values for each scaling technique. The standard IG shows a right-skewed distribution, I$\text{G}_{min}$ displays a more gradual decline across the 0-0.4 range expanding medium attributions, while I$\text{G}_{max}$ concentrates values near zero, highlighting only the strongest features.
The minimum log scale (0.6) enhances feature visibility by better distributing attribution values in the mid-range while preserving important structures.}
\label{fig:ig_scaling} 
\end{figure}

The scale factor $s \in [1,2]$ is drawn from a log-uniform distribution to obtain more evenly spread values across orders of magnitude, avoiding bias toward larger values that a linear uniform distribution would produce.
This range was selected through empirical validation: scaling factors above 2 resulted in excessive feature enhancement that left some samples with no identifiable important pixels after normalisation, while factors below 1 introduced excessive noise by emphasising less discriminative regions, thereby degrading the quality of the attribution guidance.
Figure~\ref{fig:ig_scaling} demonstrates how different scaling techniques affect the distribution of normalised IG values.
The integrated gradients are overlaid with probability $p$ to the whole image to ensure that the model also sees unaltered images and learns how to classify them.
When the image is overlaid with the integrated gradients the intensity of all pixels is first halved and then the intensity of important pixels is increased.
This targeted emphasis helps the model prioritise impactful features, improving both efficiency and interpretability.
The final training objective when combining KD, AT, and IG augmentation remains as in Equation \ref{eqn:totalloss} with IG influencing the learning process through the modified input data rather than an additional loss term.

The approach is designed to be scalable to larger datasets like ImageNet.
While the precomputation time increases linearly with dataset size, it remains a one-time cost that significantly reduces the overall training time.
Our experiments on ImageNet subsets (Section~\ref{sec:imagenet}) demonstrate the effectiveness of the approach on more complex data distributions, suggesting its viability for full ImageNet-scale models.

\subsection{Use of Artificial Intelligence}
A large language model (Claude 3.7 Sonnet by Anthropic) was used in the drafting of some sections of this manuscript.
After the initial training of our models and analysis of results, we used Claude to assist in organising and articulating our findings, particularly in the Methods and Results sections.
All AI-generated text was thoroughly reviewed, edited, and verified by the authors to ensure accuracy and alignment with our research findings.
All data analysis, figures, and technical content were produced directly by the authors without AI assistance.

\section{Experiments}
\label{sec:experiments}

\subsection{Data}
\subsubsection{Training and Testing Sets}
For our experiments, we used the standard CIFAR-10 dataset which consists of 60,000 $32 \times 32$ colour images divided into 10 classes with 6,000 images per class.
Following the standard protocol, we used 50,000 images for training and 10,000 for testing, maintaining the original balanced class distribution.
For hyperparameter optimisation, we split CIFAR-10’s standard training set (50,000 images) into training (40,000 images, 80\%) and validation (10,000 images, 20\%) subsets. 
We used this validation split to determine optimal values for knowledge distillation hyperparameters ($\alpha$, $T$), integrated gradients overlay probability ($p$), and attention transfer weight ($\gamma$).
Once optimal hyperparameters were identified, we retrained all models using the complete 50,000-image training set and evaluated final performance on the separate 10,000-image test set.

The CIFAR-10 dataset was selected for its moderate complexity, balanced classes, and widespread use as a benchmark in image classification tasks.
Its compact image size ($32 \times 32$) also allows for efficient experimentation while still presenting meaningful classification challenges that benefit from the subtle feature distinctions that our IG-enhanced approach aims to capture.

\subsubsection{Validation Set from ImageNet}
\label{sec:imagenet}
To assess the generalisation capabilities of our models beyond their training domain, we created a validation set from ImageNet classes that correspond to CIFAR-10 categories.
This pairing of CIFAR-10 with ImageNet serves multiple purposes: (1) it evaluates performance on more diverse intra-class variations and (2) it assesses whether the feature importance mechanisms learned on simpler data generalise to more complex examples.

Our validation approach involved several carefully considered steps.
First, we identified ImageNet classes that semantically align with each CIFAR-10 category.
For example, ImageNet classes such as `Airliner', `Warplane', and `Airship' were mapped to the CIFAR-10 `Plane' class.
To isolate the effect of model compression from potential teacher model errors, we performed a preliminary evaluation using the teacher model and retained only those samples correctly classified by the teacher.
This filtration process ensures that performance degradation observed in compressed models can be attributed specifically to compression effects rather than inherent limitations of the teacher model on challenging samples.
Out of the 5250 total ImageNet samples in our validation set, the teacher model correctly predicted 3537 images (67.37\% accuracy), with the remaining 1713 incorrectly predicted images being excluded from our analysis.

Table S13 (Online Resource 1) provides the complete mapping between CIFAR-10 classes and their corresponding ImageNet categories.
Notably, two CIFAR-10 classes (`Deer' and `Horse') did not have direct one-to-one correspondences in ImageNet, reflecting the taxonomic differences between the datasets.

\subsection{Hyperparameter Optimisation}
We conducted an extensive grid search to identify optimal hyperparameters for knowledge distillation, integrated gradients, and attention transfer.
Our experiments maintained a consistent architectural configuration with a compression factor of 4.1, reducing the model from 2.2M to 543K parameters.
To ensure statistical reliability, each configuration underwent 10 independent training runs long enough to account for initialisation variance.
Table~\ref{tab:hyperparam_ranges} presents the complete search space for all hyperparameters, with ranges selected based on preliminary experiments and established literature values. Note that we, in our ablation study, also test all combinations of KD, IG, and AT, including combinations corresponding to the weights being zero.

\subsubsection{Knowledge Distillation Hyperparameter Search}
The knowledge distillation optimisation focused on two key hyperparameters: the distillation weight $\alpha$ and temperature $T$.
The selected ranges for $\alpha$, as shown in Table \ref{tab:hyperparam_ranges}, were chosen to examine both subtle and strong influences of the guidance from the teacher model.
Similarly, the temperature values were selected to investigate varying degrees of softness in probability distributions.

\subsubsection{Integrated Gradients Optimisation}
Building upon the knowledge distillation framework, we investigated the impact of overlay probability $p$ for integrated gradients-based data augmentation.
This hyperparameter governs how frequently IG maps are applied to training images, requiring careful calibration to balance attribution information with the preservation of original image characteristics.

\subsubsection{Attention Transfer Optimisation}
The attention transfer investigation centred on the weight hyperparameter $\gamma$, which determines the contribution of attention map alignment in the loss function.
As detailed in Table \ref{tab:hyperparam_ranges}, we examined a broad range of values to understand the full spectrum of attention transfer influence during student model training.

\begin{table}[t!]
\centering
\caption{Hyperparameter search space for model compression.
  The table presents the explored ranges for KD, IG, and AT hyperparameters.
  Each hyperparameter range was selected based on preliminary experiments and literature review to balance model performance and training stability.
  $T$ represents the temperature for softening probability distributions, $\alpha$ represents the knowledge distillation loss weight, $p$ represents the overlay probability for integrated gradients, and $\gamma$ represents the attention transfer weight coefficient.}
  \begin{tabular}{ccc}
\hline
Component & Hyperparameter & Search Range \\
\hline
\multirow{2}{*}{KD} & $T$ & \{1.5, 2, 2.5, 3, 4\} \\
 & $\alpha$ & \{0.0005, 0.005, 0.01, 0.025, 0.05, 0.075, 0.09, 0.1, 0.25\} \\
IG & $p$ & \{0.5, 0.25, 0.2, 0.15, 0.1, 0.09\} \\
AT & $\gamma$ & \{0.9, 0.8, 0.75, 0.7, 0.6, 0.5, 0.4, 0.3, 0.25, 0.2, 0.1\} \\
\hline
\end{tabular}
\label{tab:hyperparam_ranges}
\end{table}

\subsection{Ablation Study}
To isolate the contributions of individual components, we conducted a comprehensive ablation study testing various combinations of KD, IG, and AT.
Each configuration used the optimal hyperparameters identified in the previous section and was evaluated over 10 independent runs to ensure robust performance assessment.
For this study, we used 100\% of the standard training data (50,000 images) for each run, with variation coming only from random weight initialisation and training processes.
The configurations tested included standalone approaches (KD alone, IG alone, AT alone) as well as combinations (KD \& IG, KD \& AT, IG \& AT, and KD \& IG \& AT), allowing us to assess both individual and synergistic effects between components.

\subsection{Compression Factor Analysis}
We systematically evaluated the trade-off between model size and performance by testing student models with compression factors ranging from 2.2 (relatively modest compression) to 1121.7 (extreme compression).
For moderate compression factors (2.2x-12.04x), we conducted 10 independent training runs per configuration to ensure robustness, while more extreme compression factors were evaluated with 3 training runs each.
Unlike our previous experiments which reported results from the best of multiple runs, this analysis reports the average performance to better represent the expected outcomes in practical deployment scenarios.
This methodological choice eliminated potential selection bias that might obscure the true relationship between compression and accuracy degradation.

The compression levels were achieved through progressive layer removal from the MobileNetV2 architecture, maintaining the early feature extraction layers while systematically reducing the network depth.
This approach creates a controlled experiment where the fundamental architectural characteristics remain consistent while the model capacity is progressively reduced.
Additionally, we measured both training time per epoch and inference time across all three GPU configurations (RTX 3060 Ti, RTX 3090, and RTX A5000) to evaluate how computational efficiency scales with model compression under different hardware environments.
These comprehensive measurements provide insights into the practical deployment considerations beyond accuracy metrics alone.

\subsection{Monte Carlo Simulation}
To rigorously assess the statistical robustness of our approach, we conducted Monte Carlo simulations comprising 60 independent runs for each configuration (Student baseline, KD, KD \& IG, and KD \& IG \& AT).
Unlike the ablation study, each Monte Carlo run used a randomly selected 80\% subset of the training data (40,000 images) while maintaining evaluation on the full test set (10,000 images).
This methodology reveals the distribution of potential outcomes rather than single point estimates, assesses model robustness to variations in training data, provides statistical confidence intervals for performance metrics, and eliminates potential selection bias from fortuitous initialisations or data splits.
The stochastic nature of deep learning models—arising from random weight initialisation, mini-batch selection, and optimisation dynamics—means that identical architectures and hyperparameters can yield different results across training runs.
Our Monte Carlo approach captures this inherent variability, enabling a more comprehensive assessment of the performance characteristics of each configuration.

The choice of 60 runs represents a carefully considered balance between statistical power and computational feasibility.
Our analysis indicates that reducing to 50 runs would widen the 95\% confidence interval from ±0.15\% to ±0.17\% (a 13\% increase in uncertainty), while requiring approximately 17\% less computation time.
Conversely, increasing to 70 runs would narrow the confidence interval to ±0.14\% (a 7\% improvement in precision) but require 17\% more computational resources.
Given diminishing returns in statistical precision beyond 60 runs and our computational constraints, this run count provides an optimal balance between robust statistical analysis and practical implementation.
A detailed analysis of the relationship between run count and confidence intervals for the KD \& IG configuration (4.12 compression factor) is provided in Table S8 (Online Resource 1).

\subsection{ImageNet Subset Evaluation}
Following the creation of the ImageNet validation subset described in Section \ref{sec:imagenet}, we evaluated all model configurations on these diverse images.
This cross-dataset evaluation serves as a strong test of generalisation capabilities, assessing whether the knowledge transferred from teacher to student extends beyond the specific characteristics of the training data.
The evaluation used the best-performing model from each configuration, applying it directly to the ImageNet subset without any fine-tuning or domain adaptation.
To address the resolution discrepancy between ImageNet's standard $224 \times 224$ pixel images and CIFAR-10's $32 \times 32$ pixel format, we utilised a downsampled version of ImageNet at $32 \times 32$ resolution, ensuring compatibility with our model architectures without requiring structural modifications.

\subsection{Computational Infrastructure}
All experiments were conducted using an NVIDIA GeForce RTX 3090 GPU (24GB VRAM), AMD Ryzen Threadripper 1950X CPU (16 cores/32 threads), and 96GB system RAM. 
With this hardware configuration, training a student model for 100 epochs required approximately 21 minutes, while precomputing integrated gradients for the entire CIFAR-10 training set took approximately 2 hours. 
The Monte Carlo simulations with 60 runs per configuration required approximately 84 GPU-hours in total. 
For the compression factor analysis, we utilised additional server configurations: an NVIDIA GeForce RTX 3060 Ti (8GB VRAM), AMD Ryzen 5 5600X CPU (6 cores/12 threads), and 16GB system RAM; and an NVIDIA RTX A5000 (24GB VRAM), AMD Ryzen Threadripper PRO 5955WX CPU (16 cores/32 threads), and 504GB system RAM. 
This approach allowed us to evaluate how inference and training times varied across different hardware environments as compression factors increased. 
This distributed infrastructure provided sufficient computational capacity to maintain consistent experimental conditions across all evaluations.

\section{Results and Discussion}\label{sec:results}

\subsection{Hyperparameter Optimisation Results}
Our systematic grid search across multiple hyperparameters revealed optimal configurations for knowledge distillation, integrated gradients, and attention transfer components.
These findings establish the foundation for our subsequent model compression experiments and are included in Online Resource 1.

\subsubsection{Knowledge Distillation Hyperparameter Optimisation}
The results of our KD hyperparameter optimisation revealed an optimal configuration of $\alpha$ = 0.01 and $T$ = 2.5, as summarised in Table S10 (Online Resource 1) and visualised in Figure S2 (Online Resource 1).
The surface plot demonstrates that lower $\alpha$ values (0.01-0.05) consistently preserve model performance during knowledge distillation.
This finding suggests that subtle influences from the soft targets of the teacher model provide more effective guidance than stronger distillation weights.
The optimal temperature of $T$ = 2.5 indicates that moderate softening of probability distributions strikes an effective balance between preserving class relationships and maintaining sufficient categorical distinction.

\subsubsection{Integrated Gradients Optimisation}
Building upon the optimal KD hyperparameters, our investigation of IG overlay probabilities revealed 0.1 as the optimal value, achieving 92.6\% accuracy.
See Table S11 for details (Online Resource 1).
This relatively sparse application of integrated gradients effectively balances feature emphasis with model generalisation.
Higher probabilities (0.25, 0.5) demonstrated performance degradation, suggesting that excessive attribution information may cause the model to overemphasise specific features at the expense of learning diverse representations.
Conversely, reducing the probability below 0.1 resulted in insufficient guidance about important features, as evidenced by the decline in performance at $p$ = 0.09.

\subsubsection{AT Hyperparameter Optimisation}
The optimisation of the attention transfer weight $\gamma$ yielded an optimal value of 0.8, achieving 92.4\% accuracy, see Table S12 (Online Resource 1) for details.
This relatively high weighting of attention map alignment in the loss function demonstrates the significance of attention transfer in guiding student model learning.
The performance curve exhibits a clear peak at $\gamma$ = 0.8, with notable degradation both above and below this value, indicating a sensitive optimum that effectively balances attention-based and standard classification objectives.

\subsubsection{Attention Map Analysis}
Visual inspection of attention maps in Figure S1 (Online Resource 1) reveals that student models consistently display more concentrated attention patterns than the teacher model, despite low MSE values (0.0008-0.0054).
This intensity disparity occurs because MSE measures relative attention patterns rather than absolute values, with student models focusing attention more intensely on fewer pixels.
The automobile class consistently shows higher MSE values (0.0045-0.0054) across all methods, suggesting this class presents particular challenges for attention transfer due to its complex feature structure and diverse orientations.

The KD \& IG \& AT approach shows more consistent performance across all classes, particularly improving on traditionally difficult classes like ship (reducing MSE from 0.0027 to 0.0019 compared to AT alone).
This demonstrates that while MSE provides a useful quantitative metric, the qualitative aspects of attention distribution—such as focus area consistency and feature highlighting—also play important roles in effective knowledge transfer.

\subsection{Ablation Study}
Table~\ref{table:ablation} presents the accuracies of various configurations for the student model with 4.12 compression factor.
The teacher model achieves 93.9\%, while the student model without distillation reaches 91.6\%.
These results are based on 10 independent runs using the complete training dataset (50,000 images), which accounts for the generally higher accuracies compared to our Monte Carlo simulations that used only 80\% of the training data.
To establish statistical significance, we conducted paired t-tests comparing each approach to the student baseline.
The paired test design accounts for the dependency between observations, as all configurations were evaluated on the same test sets.

\begin{table*}[bt]
\centering
\caption{Comparison of testing accuracies across different methods for the student model with 4.1 compression factor.
$\Delta$Acc. is the difference between the testing accuracy of the teacher and the highest testing accuracy of the student model.
Statistical significance was assessed using paired t-tests against the student baseline across 10 independent runs using 100\% of the training data.}
\label{table:ablation}
\begin{tabular}{p{2.3cm}p{.88cm}p{.7cm}p{.7cm}p{.7cm}cp{.75cm}p{1cm}}
\hline
\multirow{2}{*}{Method} & \multicolumn{5}{c}{Accuracy (\%)} & \multirow{2}{*}{t-stat} & \multirow{2}{*}{p-value} \\
 & $\Delta$Acc. & Max & Min & Mean & Std. Dev. & & \\
\hline
Teacher & - & 93.91 & - & - & - & - & - \\
Student & -2.41 & 91.50 & 90.50 & 91.18 & 0.29 & - & - \\
KD & -2.29 & 92.29 & 90.88 & 91.62 & 0.41 & 2.57 & 0.030 \\
KD \& IG & \textbf{-1.89} & \textbf{92.58} & \textbf{91.58} & \textbf{92.02} & 0.32 & 6.52 & $<$0.001 \\
KD \& AT & -2.16 & 92.20 & 91.39 & 91.75 & 0.27 & 3.57 & 0.006 \\
KD \& IG \& AT & -2.05 & 92.42 & 91.51 & 91.86 & 0.29 & \textbf{6.91} & \textbf{$<$0.001} \\
IG \& AT & -2.45 & 91.84 & 91.05 & 91.46 & 0.29 & 1.97 & 0.080 \\
AT & -2.75 & 91.58 & 90.62 & 91.16 & \textbf{0.23} & -0.14 & 0.895 \\
IG & -2.56 & 92.01 & 90.85 & 91.35 & 0.31 & 0.88 & 0.404 \\
\hline
\end{tabular}
\end{table*}

Applying KD alone improves accuracy to 92.3\% ($p = 0.030$), representing a substantial gain of 0.8 percentage points over the compressed model baseline.
Among all configurations, KD combined with IG achieves the highest accuracy of 92.6\% ($p < 0.001$), with a $\Delta$Acc of -1.89\% relative to the teacher, and a statistically significant improvement of 1.1 percentage points over the student baseline.
This configuration demonstrates that IG enhances distillation by guiding the student to focus on critical features, achieving the highest relative improvement of 44.8\%.

The KD \& IG \& AT combination yields 92.42\% accuracy ($p < 0.001$), closely trailing KD \& IG.
Interestingly, while this combined approach was expected to produce the best results by leveraging all three techniques, the marginal decrease in performance compared to KD \& IG suggests potential interaction effects or slight overfitting when all mechanisms are employed simultaneously.
This may be due to competing optimisation objectives between attention transfer and integrated gradients, where the focus of AT on spatial attention patterns could occasionally conflict with the emphasis of IG on feature-level attributions.
Standalone configurations of IG and AT produce lower accuracies of 92.01\% and 91.6\% respectively, while their combination (IG \& AT) achieves 91.8\%. 
Statistical analysis shows that KD \& IG ($p < 0.001$), KD \& AT ($p = 0.006$), and KD \& IG \& AT ($p < 0.001$) provide significant improvements over the baseline, while IG \& AT ($p = 0.080$), IG ($p = 0.404$), and AT ($p = 0.895$) do not reach statistical significance at the conventional $p < 0.05$ threshold.

This ablation study confirms that KD serves as the foundation of our compression framework, with IG providing significant complementary benefits.
The KD \& IG configuration emerges as the most effective, demonstrating that IG enhances feature-level alignment between teacher and student, resulting in superior accuracy and interpretability.
These results are particularly notable in the context of edge device deployment, where the 4.1x reduction in model size translates to proportional decreases in memory requirements and inference time, with only a 1.9 percentage point accuracy drop from the teacher model.

\subsection{Compression Factor Analysis}

\begin{figure*}[ht]
    \centering
    \includegraphics[width=\textwidth]{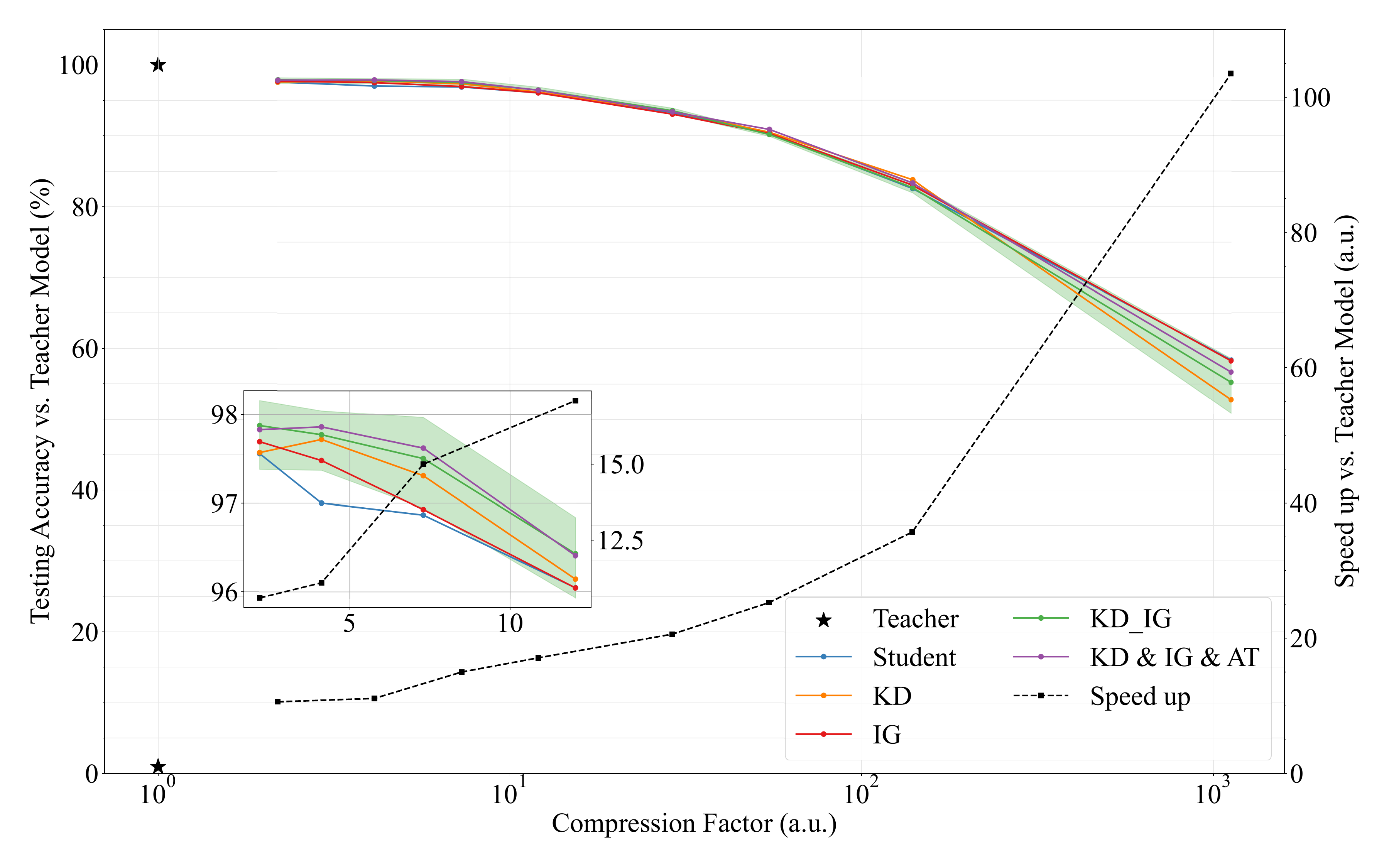}
    \caption{Testing accuracy (solid lines, left axis) and inference speed-up (dashed line, right axis) as functions of compression factor for different model configurations. 
    The main graph shows the performance-compression trade-off across the full range of compression factors (1x to 1122x), while the inset provides a detailed view of the moderate compression range (2.2x to 12x). 
    KD \& IG maintains a consistent performance advantage in this critical range while delivering computational speed-ups that exceed the compression ratio. 
    Note that all data points represent models trained only once, so stochastic fluctuations are present.}
    \label{fig:delta_acc_vs_compression}
\end{figure*}

Our analysis reveals clear patterns in how increasing compression affects model performance across different training configurations, as illustrated in Figure~\ref{fig:delta_acc_vs_compression}. 
The performance-compression relationship demonstrates both the capabilities and limitations of our approach, with several distinct operational ranges emerging from the data.
In the moderate compression range (2.2× to 12×), both KD and KD \& IG demonstrate remarkable stability, maintaining accuracies above 96\% of the teacher's performance even at 12× compression—a significant achievement considering the model size reduction to just 8.3\% of the original architecture.
The detailed view in the inset figure highlights that at 4.1× compression, our primary experimental configuration, KD \& IG achieves approximately 98.6\% of the teacher model's accuracy while substantially reducing computational demands.
Beyond 28× compression, all configurations experience accelerated performance degradation, as shown by the steeper decline in the accuracy curves. 
The differences between techniques become less pronounced, with the performance curves converging. 
This pattern suggests that at extreme compression levels, the fundamental limitations of model capacity overshadow the benefits of sophisticated knowledge transfer approaches.
This becomes particularly evident at our highest tested compression ratio (1122×), where all configurations experience a significant drop to around 55-60\% of the teacher model's accuracy. At this extreme compression level, the performance differences between Student, KD, IG, and KD \& IG models become minimal, indicating that the benefits of knowledge transfer are largely neutralised when model capacity is severely constrained.
Importantly, KD \& IG maintains a consistent performance advantage throughout most of the compression spectrum, particularly in the 4-30× compression range—the most relevant range for practical edge deployments. 
Figure~\ref{fig:delta_acc_vs_compression} also illustrates the computational efficiency gains (dashed line, right axis), showing that speed-up factors exceed the compression ratios, with our 4.1× compressed model achieving approximately 10× inference speedup, and our most compressed model (1122×) achieving over 100× speedup.
These results demonstrate that our method provides reliable and predictable performance-size trade-offs that can be effectively tailored to specific deployment requirements. 
Detailed computational efficiency and inference time analyses are provided in Figures S5–S6 and  Table S9 (Online Resource 1), with full performance metrics available in Table S15 (Online Resource 1).

\subsection{Monte Carlo Simulation} \label{sec:montecarlo_results}
\begin{figure}[]
  \centering 
  \includegraphics[width=\textwidth]{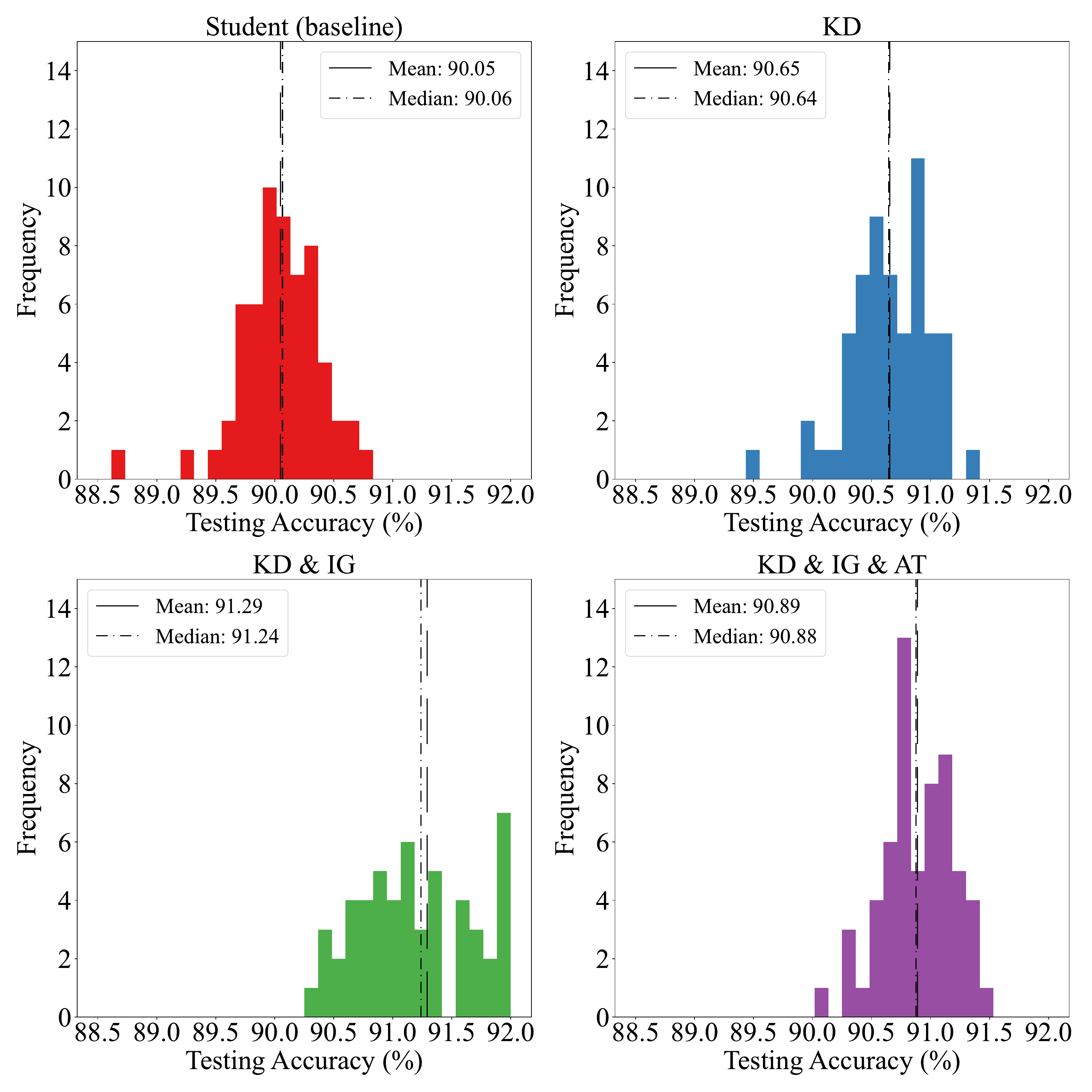} 
  \caption{Distributions of testing accuracies from Monte Carlo simulations across various methods.
  The histograms depict the performance variability of four configurations: Student (baseline), KD, KD \& IG, and KD \& IG \& AT.
  The mean and median testing accuracies are indicated for each method, showcasing the influence of knowledge distillation, integrated gradients, and attention transfer on model performance.} 
  \label{fig:montecarlo} 
\end{figure}

Figure~\ref{fig:montecarlo} illustrates the distributions of testing accuracies from 60 Monte Carlo simulations for four configurations, providing robust statistical evidence of performance differences.
The baseline student model achieves a mean accuracy of 90.05\% (median 90.06\%), representing the performance of our compressed architecture without advanced knowledge transfer trained on only 80\% of randomly picked images from the training set.
KD improves the mean accuracy to 90.65\% (median 90.64\%), demonstrating the value of soft targets in guiding student learning.
Most notably, KD with IG yields a mean accuracy of 91.29\% (median 91.24\%), confirming it as the highest-performing configuration with a statistically significant improvement of 1.24 percentage points over the student baseline ($p < 0.001$).
The KD \& IG \& AT configuration achieves a mean accuracy of 90.89\% (median 90.88\%), performing better than KD alone but not matching KD \& IG.

The paired t-tests (Table~\ref{tab:statistical_significance}) confirm the statistical significance of the observed improvements, with all distillation approaches showing significant gains over the student baseline ($p < 0.001$).
The KD \& IG configuration shows the highest t-statistic (14.80), indicating the most robust improvement.

\begin{table*}[h]
    \centering
    \caption{Statistical analysis of the testing accuracies obtained from the Monte Carlo simulation results comparing different approaches for the student model with 4.1 compression factor.
    Paired t-tests were conducted against the Student baseline using data from 60 independent runs, each using 80\% of the training data.
    }
    \label{tab:statistical_significance}
    \begin{tabular}{ccccc}\hline
        Method & Mean & Std Dev & t-statistic & p-value \\
        \hline
        Student & 90.05\% & 0.343\% & - & - \\
        KD & 90.65\% & 0.337\% & 9.74 & $<$0.001 \\
        KD \& IG & 91.29\% & 0.553\% & 14.80 & $<$0.001 \\
        KD \& IG \& AT & 90.89\% & 0.297\% & 14.32 & $<$0.001 \\
        \hline
    \end{tabular}
\end{table*}

Interestingly, the KD \& IG configuration shows higher variance (std. dev. 0.553\%) than other approaches, suggesting that while it achieves the highest mean performance, it may be more sensitive to data subset selection and initialisation conditions.
This characteristic indicates that in deployment scenarios where consistent performance is prioritised over maximum accuracy, the more stable KD \& IG \& AT approach (std. dev. 0.297\%) might be preferable despite its slightly lower mean accuracy.

These results confirm our approach delivers consistent improvements across different training conditions, independent of initialisation or dataset variations.

\subsection{ImageNet Subset Evaluation}
Our evaluation on the curated ImageNet subset demonstrates the broad generalisation capabilities of our approach, as shown in Figure S4 (Online Resource 1).
The KD \& IG configuration achieves 85.7\% accuracy on ImageNet, significantly outperforming the 83.8\% accuracy of the baseline student model, while maintaining strong performance relative to the 100\% baseline of the teacher model.
This performance advantage is particularly noteworthy given the substantial domain shift between training and evaluation conditions.
While trained exclusively on CIFAR-10's $32 \times 32$ pixel images, our models maintain robust performance when evaluating ImageNet's more challenging $224 \times 224$ pixel images, which exhibit greater intra-class variation and complexity.
The improved accuracy of KD \& IG over standalone KD (85.0\%) and IG (85.1\%) approaches suggests that our combined methodology helps models learn more robust and transferable features.
The consistent superior performance of KD \& IG across both CIFAR-10 (92.6\%) and ImageNet (85.7\%) datasets, as detailed in Table~\ref{tab:imagenet_eval}, indicates that our approach enhances the ability of the model to identify and leverage class-relevant features rather than dataset-specific characteristics.
This cross-dataset generalisation capability is crucial for real-world applications where deployment conditions may differ significantly from training scenarios.

\begin{table*}[t!]
\centering
\caption{Performance comparison of different knowledge distillation configurations on CIFAR-10 and ImageNet subsets}
\label{tab:imagenet_eval}
\begin{tabular}{lcc}
\hline
Model Configuration & CIFAR-10 (\%) & ImageNet subset (\%) \\
\hline
Teacher & 93.91 & 100.00 \\
Student (baseline) & 91.50 & 83.78 \\
KD & 92.29 & 84.97 \\
KD \& IG & \textbf{92.58} & \textbf{85.71} \\
KD \& IG \& AT & 92.42 & 85.22 \\
IG & 92.01 & 85.07 \\
\hline
\end{tabular}
\end{table*}

\section{Conclusions}\label{sec:conclusions}
Our research presents a novel approach to model compression that effectively balances model size reduction, performance preservation, and interpretability.
Through systematic evaluation on both CIFAR-10 and ImageNet datasets, we demonstrated that combining knowledge distillation with integrated gradients significantly enhances student model performance, achieving an accuracy of 92.6\% at 4.1x compression—the highest observed in our ablation study while preserving 98.6\% of the performance of the teacher model.
The integration of IG with KD provides two key advantages over traditional compression methods.
First, it enables more effective knowledge transfer by guiding the student model to focus on the same discriminative features as the teacher, as evidenced by our attention map analysis.
Second, it maintains model interpretability through visualisation of feature attributions, a critical requirement for real-world deployments where understanding model decisions is essential.
Our extensive empirical validation, including Monte Carlo simulations and comprehensive compression factor analysis, establishes the robustness of this approach across varying degrees of compression.
The method demonstrates particular effectiveness in the practical compression range of 4-30x, making it well-suited for edge device deployment scenarios.
The computational benefits extend beyond model size reduction, with our 4.1x compressed model achieving a 11.1x speedup in inference time.
Looking ahead, this research opens new avenues for developing compression techniques that preserve not only model performance but also interpretability.
Our methodology provides a foundation for future work in creating efficient, explainable models suitable for resource-constrained environments.

\section{Statements and Declarations}

\subsection{Availability of data and material}
The CIFAR-10 dataset (\url{https://www.cs.toronto.edu/~kriz/cifar.html}) and ImageNet dataset (\url{https://image-net.org/}) used in this study are publicly available. The processed ImageNet subset described in Section~\ref{sec:imagenet} can be accessed through our code repository.

\subsection{Competing Interests}
The authors have no financial or non-financial interests that are directly or indirectly related to the work submitted for publication.

\subsection{Funding}
This research was supported by the National Science and Technology Council (NSTC) of Taiwan under Grant No. NSTC 112-2314-B-006-079 (Development of a COVID-19 forecasting model based on synthetic individual data) and Grant No. NSTC 113-2314-B-006-069 (Forecasting of COVID-19 based on synthetic individual data). This research was supported in part by Higher Education Sprout Project, Ministry of Education to the Headquarters of University Advancement at National Cheng Kung University (NCKU).

\subsection{Acknowledgements}
We acknowledge the National Cheng Kung University's Mechanical Engineering Department for providing the computational resources necessary to conduct our experiments. We also thank Paul Tsai for providing inspiration through his former thesis work.

\subsection{Authors’ Contribution Statement}
Author contribution using the CRediT taxonomy: 
Conceptualisation: TN and JC; Data curation: DH; Formal analysis: JC and DH; Methodology: JC; Investigation: DH; Software: DH; Verification: DH and JC and TN; Visualisation: DH, JC, and TN; Writing - original draft preparation: DH; Writing - review and editing: JC and TN; Funding acquisition: TN; Project administration: TN; Resources: TN; Supervision: JC and TN.

\subsection{Code Availability}
Our models and code are available in the following repository: \url{https://github.com/nordlinglab/ModelCompression-IntegratedGradients}.

\bibliography{References}

\end{document}

\title[KD-IG]{Model compression using knowledge distillation with integrated gradients--ESM}

\author[1]{\fnm{David E.} \sur{Hernandez}, \dgr{B.S.}}\email{david.hernandez@nordlinglab.org (0009-0004-4392-6927)}

\author[1]{\pfx{Dr} \fnm{Jose Ramon} \sur{Chang}, \dgr{MSc, PhD}}\email{jose.chang@nordlinglab.org (0000-0001-5587-7828)}

\author*[1]{\pfx{Dr} \fnm{Torbj{\"o}rn E. M.} \sur{Nordling}, \dgr{MSc, PhD}}\email{torbj{\"o}rn.nordling@nordlinglab.org (0000-0003-4867-6707)}

\affil[1]{\orgdiv{Department of Mechanical Engineering}, \orgname{National Cheng Kung University}, \orgaddress{\street{No. 1 University Rd.}, \city{Tainan} \postcode{701}, \country{Taiwan}}}

\maketitle

This document provides supplementary information for the manuscript ``Model compression using knowledge distillation with integrated gradients." 
It contains comprehensive analyses and extended results that support our main findings while offering deeper technical insights. 
The supplement begins with a detailed literature review examining model compression techniques across multiple datasets, providing statistical comparisons of accuracy-compression tradeoffs. 
We then present detailed architectural specifications of our teacher and student models across different compression factors (2.2× to 1122×), followed by attention map analyses that visualise how our approach maintains feature focus. 
The document includes comprehensive hyperparameter optimisation results for knowledge distillation, integrated gradients overlay probabilities, and attention transfer weights. 
Statistical validation through Lilliefors tests and Monte Carlo simulations demonstrates the robustness of our approach. 
We provide cross-dataset evaluations using ImageNet subsets to demonstrate generalisation capabilities beyond CIFAR-10, visualisations of integrated gradients on a higher-complexity dataset, and detailed computational performance analyses quantifying inference speedups. 
Complete tables of testing accuracies across different compression configurations are included to facilitate comparison with future research.

\section{Supplementary}

\subsection{Literature Review Details}
Our comprehensive literature review examined model compression techniques across multiple datasets and architectures.
Tables~\ref{tab:paper_results} and \ref{tab:paper_results2} present the testing accuracies and compression factors achieved by various student models compared to their teacher models.
These results demonstrate the evolution of knowledge distillation techniques and their effectiveness across different datasets.
Tables~\ref{tab:paper_results3} and \ref{tab:paper_results4} provide the corresponding model parameters and architectural details for each configuration, enabling detailed analysis of how model structure influences compression outcomes.

The results are organised to show both the highest accuracy and highest compression achievements for each study, with testing accuracies reported unless specifically noted with an asterisk (*) for training accuracies.
As shown in Table 2 (main article), different datasets exhibited varying levels of compressibility, with simpler datasets like MNIST generally achieving higher compression factors while maintaining accuracy.

The relationship between model size and performance preservation, visualised in Figure 1 (main article), reveals that similar architectures often achieved notably different compression factors.
This suggests that the distillation process itself, rather than just architectural choices, plays a crucial role in determining final model efficiency.

\begin{table}
\centering
\caption{
Results comparing teacher and student model accuracies across various datasets (Part 1). 
Only the highest accuracy and highest compression results for each study were included. 
All accuracies are assumed to be testing accuracies unless noted with an asterisk (*), which indicates training accuracies, ``Baseline" refers to the accuracy of the student model before any knowledge distillation has been applied, while ``Distilled" denotes the accuracy after applying the distillation process.
``Rel. $\Delta$Acc. means the Relative $\Delta$Accuracy between the Teacher, Student baseline, and Student distilled accuracies. The corresponding model parameters are detailed in Table~\ref{tab:paper_results3}.
}
\label{tab:paper_results}

\begin{tabular}{p{1cm}p{.5cm}p{1cm}p{1cm}p{1cm}p{1cm}p{1.5cm}p{2.5cm}}
\hline
Teacher  & \multirow{2}{*}{CF}  & \multicolumn{2}{c}{Student Acc. (\%)} & \multirow{2}{*}{$\Delta$Acc.} & Rel. & \multirow{2}{*}{Dataset}  & \multirow{2}{*}{Article}            \\
  Acc. (\%)                  &                      & Baseline & Distilled                  &                               &   $\Delta$Acc. (\%)    &                           &                                     \\ 
\hline
\multirow{2}{*}{99.48*}            & 15.7                 & -        & 99.28*                     & -0.20                         & -                               & \multirow{2}{*}{MNIST}    & \multirow{6}{*}{\citet{wang2019private}}    \\
                                   & 32.23                & -        & 98.94*                     & -0.54                         & -                               &                           &                                     \\
\multirow{2}{*}{96.36*}            & 20.14                & -        & 95.39*                     & -0.97                         & -                               & \multirow{2}{*}{SVHN}     &                                     \\
                                   & 45.25                & -        & 94.49*                     & -1.87                         & -                               &                           &                                     \\
\multirow{2}{*}{86.35*}            & 6.00                 & -        & 84.57*                     & -1.78                         & -                               & \multirow{2}{*}{CIFAR10}  &                                     \\
                                   & 20.80                & -        & 82.14*                     & 4.21                          & -                               &                           &                                     \\
                                   & \multicolumn{1}{l}{} &          &                            & \multicolumn{1}{l}{}          & \multicolumn{1}{l}{}            &                           &                                     \\
96.14                              & 3.35                 & 94.48    & 95.93                      & -0.21                         & 87.35                           & SVHN                      & \multirow{4}{*}{\citet{su2022stkd}}         \\
95.06                              & 3.26                 & 93.50    & 94.69                      & -0.37                         & 76.28                           & CIFAR10                   &                                     \\
76.35                              & 3.26                 & 72.45    & 75.15                      & -1.20                         & 69.23                           & CIFAR100                  &                                     \\
78.46                              & 2.35                 & 76.41    & 78.09                      & -0.37                         & 81.95                           & ImageNet                  &                                     \\
                                   & \multicolumn{1}{l}{} &          &                            & \multicolumn{1}{l}{}          & \multicolumn{1}{l}{}            &                           &                                     \\
76.02                              & 1.90                 & 74.13    & 75.01                      & -1.01                         & 46.56                           & \multirow{2}{*}{CIFAR100} & \multirow{3}{*}{\citet{chen2019knowledge}} \\
77.94                              & 2.06                 & 72.99    & 77.20                      & -0.74                         & 85.05                           &                           &                                     \\
78.31                              & 36.67                & 68.01    & 71.82                      & -6.49                         & 36.99                           & ImageNet                  &                                     \\
                                   & \multicolumn{1}{l}{} &          &                            & \multicolumn{1}{l}{}          & \multicolumn{1}{l}{}            &                           &                                     \\
75.61                              & 3.95                 & 71.98    & 76.62                      & +1.01                         & 127.82                          & \multirow{2}{*}{CIFAR100} & \multirow{2}{*}{\citet{hossain2024purf}}    \\
79.42                              & 6.04                     & 72.50    & 76.39                      & -3.03                         & 56.21                           &                           &                                     \\
                                   & \multicolumn{1}{l}{} &          &                            & \multicolumn{1}{l}{}          & \multicolumn{1}{l}{}            &                           &                                     \\
  98.91                              & 3.88                 & 98.65        & 98.20                      & -0.71                         & -173.08                                & \multirow{2}{*}{MNIST}                     & \multirow{5}{*}{\citet{chen2019data}}      \\
  98.39   & 1.88 & 98.11 & 97.91 & -0.48 & -71.43 & & \\ 
95.58                              & 1.91                 & 93.92        & 92.22                      & -3.36                         & -102.41                              & CIFAR10                   &                                     \\
77.84                              & 1.91                 & 76.53        & 74.47                      & -3.37                         & -157.25                               & CIFAR100                  &                                     \\
81.59                             & 3.20                 & 80.82        & 80.03                      & -1.56                         & -102..60                               & CelebA                    &                                     \\
                                   & \multicolumn{1}{l}{} &          &                            & \multicolumn{1}{l}{}          & \multicolumn{1}{l}{}            &                           &                                     \\
89.70                              & 1.36                 & 82.60       & 88.70                      & -1.00                         & 174.29                               & CUHK03                    & \multirow{2}{*}{\citet{chen2018darkrank}}   \\
89.80     & 1.36  & 80.30    & 86.70  & -3.10   & 67.37 & Market1501 &  \\
                                   & \multicolumn{1}{l}{} &          &                            & \multicolumn{1}{l}{}          & \multicolumn{1}{l}{}            &                           &                                     \\
98.04                              & 11.00                & 97.67    & 96.48                      & -1.56                         & -321.62                         & SVHN                      & \multirow{5}{*}{\citet{choi2020data}}       \\
95.11                              & 1.90                 & 94.56    & 94.61                      & -0.50                         & 9.09                            & \multirow{2}{*}{CIFAR10}  &                                     \\
94.77                              & 11.00                & 90.97    & 86.14                      & -8.63                         & -127.10                         &                           &                                     \\
78.34                              & 1.90                 & 77.32    & 77.01                      & -1.33                         & -30.39                          & CIFAR100                  &                                     \\
66.34                              & 1.89                 & 64.78    & 63.73                      & -2.61                         & -67.31                          & Tiny-ImageNet             &                                     \\
                                   & \multicolumn{1}{l}{} &          &                            & \multicolumn{1}{l}{}          & \multicolumn{1}{l}{}            &                           &                                     \\
99.54                              & 127.02               & -        & 99.55                      & +0.01                         & -                               & MNIST                     & \multirow{7}{*}{\citet{ashok2017n2n}}       \\
95.24                              & 19.80                & -        & 95.38                      & +0.18                         & -                               & SVHN                      &                                     \\
92.05                              & 10.28                & -        & 92.35                      & +0.30                         & -                               & \multirow{2}{*}{CIFAR10}  &                                     \\
91.97                              & 20.53                & -        & 91.64                      & -0.33                         & -                               &                           &                                     \\
72.86                              & 5.02                 & -        & 70.11                      & -2.75                         & -                               & \multirow{2}{*}{CIFAR100} &                                     \\
72.22                              & 4.64                 & -        & 68.01                      & -4.21                         & -                               &                           &                                     \\
47.65                              & 3.12                 & -        & 44.71                      & -2.94                         & -                               & Caltech256                &                                     \\
\hline
\end{tabular}
\end{table}

\begin{table}
\centering
\caption{Results comparing teacher and student model accuracies across various datasets (Part 2). Only the highest accuracy and highest compression results for each study were included. All accuracies are assumed to be testing accuracies unless noted with an asterisk (*), which indicates training accuracies, ``Baseline" refers to the accuracy of the student model before any knowledge distillation has been applied, while ``Distilled" denotes the accuracy after applying the distillation process. The corresponding model parameters are detailed in Table~\ref{tab:paper_results4}.}
\label{tab:paper_results2}
\begin{tabular}{p{1cm}p{.5cm}p{1cm}p{1cm}p{1cm}p{1cm}p{1.5cm}p{2.5cm}}
\hline
Teacher  & \multirow{2}{*}{CF}  & \multicolumn{2}{c}{Student Acc. (\%)} & \multirow{2}{*}{$\Delta$Acc.} & Rel. & \multirow{2}{*}{Dataset}  & \multirow{2}{*}{Article}            \\
  Acc. (\%)                  &                      & Baseline & Distilled                  &                               &   $\Delta$Acc. (\%)    &                           &                                     \\ 
\hline
\multirow{2}{*}{79.42}             & 3.30                 & 76.35    & 79.29                      & -0.13                         & 95.76                                                    & \multirow{2}{*}{CIFAR100}     & \multirow{3}{*}{\citet{chen2022knowledge}}     \\
                                   & 2.12                 & 74.15    & 79.54                      & +0.12                         & 102.28                                                   &                               &                                                                 \\
76.26                              & -                    & 70.58    & 71.66                      & -4.60                         & 19.01                                                    & ImageNet                      &                                                                 \\
                                   & \multicolumn{1}{l}{} &          &                            & \multicolumn{1}{l}{}          & \multicolumn{1}{l}{}                                     &                               &                                                                 \\
78.92                              & 1.8                  & 78.56    & 80.46                      & +1.54                         & 527.78                                                   & \multirow{2}{*}{CIFAR100}     & \multirow{3}{*}{\citet{gou2022multilevel}}     \\
79.87                              & 52.2                 & 67.33    & 72.76                      & -7.11                         & 43.30                                                    &                               &                                                                 \\
73.11                              & 1.90                 & 69.42    & 70.92                      & -2.19                         & 40.65                                                    & ImageNet                      &                                                                 \\
                                   & \multicolumn{1}{l}{} &          &                            & \multicolumn{1}{l}{}          & \multicolumn{1}{l}{}                                     &                               &                                                                 \\
\multirow{2}{*}{95.49}             & 19.35                & -        & 94.53                      & -0.96                         & -                                                        & \multirow{2}{*}{CIFAR10}      & \multirow{3}{*}{\citet{bhardwaj2019memory}}    \\
                                   & 20.7                 & -        & 94.32                      & -1.17                         & -                                                        &                               &                                                                 \\
79.28                              & 42.94                & -        & 75.74                      & -3.54                         & -                                                        & CIFAR100                      &                                                                 \\
                                   & \multicolumn{1}{l}{} &          &                            & \multicolumn{1}{l}{}          & \multicolumn{1}{l}{}                                     &                               &                                                                 \\
87.32                              & 0.00                 & -        & 86.59                      & -0.73                         & -                                                        & CIFAR10                       & \multirow{3}{*}{\citet{blakeney2020parallel}}  \\
72.14                              & 0.00                 & -        & 70.77                      & -1.61                         & -                                                        & CIFAR100                      &                                                                 \\
65.70                              & 0.00                 & -        & 65.10                      & -0.60                         & -                                                        & ImageNet                      &                                                                 \\
                                   & \multicolumn{1}{l}{} &          &                            & \multicolumn{1}{l}{}          & \multicolumn{1}{l}{}                                     &                               &                                                                 \\
95.39                              & 1.91                 & 94.93    & 95.53                      & +0.14                         & 130.43                                                   & \multirow{2}{*}{CIFAR10}      & \multirow{5}{*}{\citet{gou2023hierarchical}}   \\
95.01                              & -                    & 91.03    & 93.58                      & -1.43                         & 64.07                                                    &                               &                                                                 \\
78.88                              & 1.67                 & 78.55    & 79.74                      & +0.86                         & 360.61                                                   & \multirow{2}{*}{CIFAR100}     &                                                                 \\
75.77                              & -                    & 70.10    & 72.98                      & -2.79                         & 50.79                                                    &                               &                                                                 \\
73.06                              & 1.90                 & 69.36    & 70.66                      & -2.40                         & 35.14                                                    & ImageNet                      &                                                                 \\
                                   & \multicolumn{1}{l}{} &          &                            & \multicolumn{1}{l}{}          & \multicolumn{1}{l}{}                                     &                               &                                                                 \\
95.89                              & 3.29                 & 94.48    & 95.83                      & -0.06                         & 95.74                                                    & SVHN                          & \multirow{5}{*}{\citet{zhao2020highlight}}     \\
94.70                              & 3.19                 & 93.68    & 94.42                      & -0.30                         & 72.55                                                    & \multirow{2}{*}{CIFAR10}      &                                                                 \\
93.43                              & 3.52                 & 91.28    & 92.50                      & -0.93                         & 56.74                                                    &                               &                                                                 \\
75.42                              & 3.19                 & 72.27    & 74.70                      & -0.72                         & 77.14                                                    & CIFAR100                      &                                                                 \\
56.51                              & 3.29                 & 50.65    & 53.59                      & -2.92                         & 50.17                                                    & Tiny-ImageNet                 &                                                                 \\
                                   & \multicolumn{1}{l}{} &          &                            & \multicolumn{1}{l}{}          & \multicolumn{1}{l}{}                                     &                               &                                                                 \\
\multirow{2}{*}{94.50}             & 8.38                 & 87.70    & 94.40                      & -0.10                         & 98.53                                                    & \multirow{2}{*}{Market1501}   & \multirow{4}{*}{\citet{xie2021model}}          \\
                                   & 15.53                & 80.90    & 92.10                      & -2.40                         & 82.35                                                    &                               &                                                                 \\
\multirow{2}{*}{86.70}             & 8.38                 & 75.30    & 86.70                      & 0.00                          & 100                                                      & \multirow{2}{*}{DukeMTMCreID} &                                                                 \\
                                   & 15.53                & 68.40    & 83.40                      & -3.30                         & 81.97                                                    &                               &                                                                 \\
\hline
\end{tabular}
\end{table}

\begin{table}
\centering
\caption{Parameters of the teacher and student models corresponding to the accuracies reported in Table~\ref{tab:paper_results}.  }
\label{tab:paper_results3}
\begin{tabular}{p{1.6cm}p{1cm}p{2cm}p{1cm}p{2cm}p{2.8cm}} 
\hline
Teacher Model             & $\#$ Params              & Student Model             & $\#$ Params            & Dataset                   & Article                                                      \\ 
\hline
-                         & \multirow{2}{*}{155.21k} & -                         & 9.88k                  & \multirow{2}{*}{MNIST}    & \multirow{6}{*}{\citet{wang2019private}}    \\
-                         &                          & -                         & 4.97k                  &                           &                                                              \\
-                         & \multirow{2}{*}{1.41M}   & -                         & 0.07M                  & \multirow{2}{*}{SVHN}     &                                                              \\
-                         &                          & -                         & 0.04M                  &                           &                                                              \\
-                         & \multirow{2}{*}{3.12M}   & -                         & 0.52M                  & \multirow{2}{*}{CIFAR10}  &                                                              \\
-                         &                          & -                         & 0.15M                  &                           &                                                              \\
                          &                          &                           &                        &                           &                                                              \\
WRN-40-1                  & 0.57M                    & WRN-16-1                  & 0.17M                  & SVHN                      & \multirow{4}{*}{\citet{su2022stkd}}         \\
\multirow{2}{*}{WRN-40-2} & \multirow{2}{*}{2.25M}   & \multirow{2}{*}{WRN-16-2} & \multirow{2}{*}{0.69M} & CIFAR10                   &                                                              \\
                          &                          &                           &                        & CIFAR100                  &                                                              \\
ResNet-152                & 60.19M                   & ResNet-50                 & 25.56M                 & ImageNet                  &                                                              \\
                          &                          &                           &                        &                           &                                                              \\
CondenseNet-86            & 0.55M                    & Smaller CondenseNet-86    & 0.29M                  & \multirow{2}{*}{CIFAR100} & \multirow{3}{*}{\citet{chen2019knowledge}}  \\
DenseNet-100              & 7.2M                     & MobileNet                 & 3.5M                   &                           &                                                              \\
ResNet-152                & 60.19M                   & MobileNet-v2              & 3.4M                   & ImageNet                  &                                                              \\
                          &                          &                           &                        &                           &                                                              \\
WRN-40-2                  & 2.25M                    & WRN-40-1                  & 0.57M                  & \multirow{2}{*}{CIFAR100} & \multirow{2}{*}{\citet{hossain2024purf}}    \\
  ResNet32x4                & 7.43M & ResNet8x4 & 1.23M &  &               \\
                          &                          &                           &                        &                           &                                                              \\
  LeNet-5                    & 62k                      & LeNet-5-HALF                         & 16k                    & \multirow{2}{*}{MNIST}                     & \multirow{5}{*}{\citet{chen2019data}}       \\
  Hinton-784-1200-1200-10 & 2.4M & Hinton-784-800-800-10 & 1.28M & \\
  \multirow{2}{*}{ResNet-34}  & \multirow{2}{*}{~21M}    & \multirow{2}{*}{ResNet-18} & \multirow{2}{*}{~11M}  & CIFAR10                   &                                                              \\
 &                          &                          &                        & CIFAR100                  &                                                              \\
AlexNet                         & 711M                     & AlexNet-Half                         & 222M                   & CelebA                    &                                                              \\
                          &                          &                           &                        &                           &                                                              \\
  \multirow{2}{*}{Inception-BN}                         & \multirow{2}{*}{10.3M}   & \multirow{2}{*}{NIN-BN}                         & \multirow{2}{*}{7.6M}  & CUHK03                    & \multirow{2}{*}{\citet{chen2018darkrank}}   \\
&  &     &      & Market1501   &                                  \\
                          &                          &                           &                        &                           &                                                              \\
WRN-40-2                  & 2.2M                     & WRN-16-1                  & 0.2M                   & SVHN                      & \multirow{5}{*}{\citet{choi2020data}}       \\
ResNet-34                 & 21.3M                    & ResNet18                  & 11.2M                  & \multirow{2}{*}{CIFAR10}  &                                                              \\
WRN-40-2                  & 2.2M                     & WRN-16-1                  & 0.2M                   &                           &                                                              \\
ResNet-34                 & 21.3M                    & ResNet18                  & 11.2M                  & CIFAR100                  &                                                              \\
ResNet-34                 & 21.4M                    & ResNet-18                 & 11.3M                  & Tiny-ImageNet             &                                                              \\
                          &                          &                           &                        &                           &                                                              \\
VGG13 & 9.4M & Layer removal \& shrinkage & 73k & MNIST & \multirow{7}{*}{\citet{ashok2017n2n}}  \\
ResNet-18   & 11.17M & Layer removal \& shrinkage & 564k  & SVHN &                      \\
ResNet-34 & 21.28M & Layer removal \& shrinkage & 2.07M & \multirow{2}{*}{CIFAR10} & \\
VGG19                     & 20.2M  & Layer removal \& shrinkage & 984k                   &                           &                                                              \\
ResNet-34   & 21.33M  & Layer removal \& shrinkage & 2.42M                  & \multirow{2}{*}{CIFAR100} &                                                              \\
ResNet-18                 & 11.22M                   & Layer removal \& shrinkage & 4.25M & & \\
ResNet-18 & 11.31M & Layer removal \& shrinkage & 3.62M  & Caltech256 & \\
\hline
\end{tabular}
\end{table}

\begin{table}
\centering
\caption{Continuation of the model parameters corresponding to the accuracies reported in Table~\ref{tab:paper_results2}.   }
\label{tab:paper_results4}
\begin{tabular}{p{1.6cm}p{1cm}p{2cm}p{1cm}p{2cm}p{2.8cm}} 
\hline
Teacher Model               & $\#$ Params             & Student Model        & $\#$ Params          & Dataset                       & Article                                                         \\ 
\hline
  \multirow{2}{*}{ResNet32x4} & \multirow{2}{*}{7.43M}                       & WRN-40-2             & 2.25M                & \multirow{2}{*}{CIFAR100}     & \multirow{3}{*}{\citet{chen2022knowledge}}     \\
 &   & ShuffleNet-v2x1.5    & 3.50M & &                   \\
-                           & -                       & ResNet-18            & 11.3M                & ImageNet                      &                                                                 \\
                            &                         &                      &                      &                               &                                                                 \\
ResNet-101                  & 42.70M                  & WRN-28-10            & 23.71M               & \multirow{2}{*}{CIFAR100}     & \multirow{3}{*}{\citet{gou2022multilevel}}     \\
ResNet-50                   & 36.54M                  & WRN-16-2             & 0.70M                &                               &                                                                 \\
ResNet-34                   & 21.33M                  & ResNet-18            & 11.22M               & ImageNet                      &                                                                 \\
                            &                         &                      &                      &                               &                                                                 \\
\multirow{2}{*}{WRN-40-4}   & \multirow{2}{*}{8.9M}   & NoNN-2S-XL           & 0.46M                & \multirow{2}{*}{CIFAR10}      & \multirow{3}{*}{\citet{bhardwaj2019memory}}    \\
                            &                         & NoNN-2S              & 0.43M                &                               &                                                                 \\
WRN-28-10                   & 36.5M                   & NoNN-2S              &                      & CIFAR100                      &                                                                 \\
                            &                         &                      &                      &                               &                                                                 \\
VGG16                       & 138.36M                 & VGG16                & 138.36M              & CIFAR10                       & \multirow{3}{*}{\citet{blakeney2020parallel}}  \\
ResNet-50                   & 25.56M                  & ResNet-50            & 25.56M               & CIFAR100                      &                                                                 \\
VGG16                       & 138.36M                 & VGG16                & 138.36M              & ImageNet                      &                                                                 \\
                            &                         &                      &                      &                               &                                                                 \\
ResNet-34                   & 21.30M                  & ResNet-18            & 11.17M               & \multirow{2}{*}{CIFAR10}      & \multirow{5}{*}{\citet{gou2023hierarchical}}   \\
ResNet-18                   & 11.17M                  & ShuffleNet-v2        &                      &                               &                                                                 \\
ResNet-101                  & 42.70M                  & ResNet-50            & 25.56M               & \multirow{2}{*}{CIFAR100}     &                                                                 \\
ResNet-18                   & 11.22M                  & ShuffleNet-v2        &                      &                               &                                                                 \\
ResNet-34                   & 21.33M                  & ResNet-18            & 11.22M               & ImageNet                      &                                                                 \\
                            &                         &                      &                      &                               &                                                                 \\
WRN-40-1                    & 0.56M                   & WRN-16-1             & 0.17M                & SVHN                          & \multirow{5}{*}{\citet{zhao2020highlight}}     \\
WRN-40-2                    & 2.20M                   & WRN-16-2             & 0.69M                & \multirow{2}{*}{CIFAR10}      &                                                                 \\
WRN-40-1                    & 0.57M                   & WRN-16-1             & 0.17M                &                               &                                                                 \\
WRN-40-2                    & 2.20M                   & WRN-16-2             & 0.69M                & CIFAR100                      &                                                                 \\
WRN-40-1                    & 0.56M                   & WRN-16-1             & 0.17M                & Tiny-ImageNet                 &                                                                 \\
                            &                         & \multicolumn{1}{l}{} & \multicolumn{1}{l}{} &                               &                                                                 \\
\multirow{2}{*}{ResNet-50}  & \multirow{2}{*}{25.32M} & Pruned0.95           & 3.02M                & \multirow{2}{*}{Market1501}   & \multirow{4}{*}{\citet{xie2021model}}          \\
                            &                         & ShuffleNet           & 1.63M                &                               &                                                                 \\
\multirow{2}{*}{ResNet-50}  & \multirow{2}{*}{25.32M} & Pruned0.95           & 3.02M                & \multirow{2}{*}{DukeMTMCreID} &                                                                 \\
                            &                         & ShuffleNet           & 1.63M                &                               &                                                                 \\
\hline
\end{tabular}
\end{table}

\subsubsection{Literature Search Methodology}

The following search strings were used with minor modifications to search Scopus, Web of Science, and Google Scholar databases for relevant literature on model compression using knowledge distillation techniques:

\begin{itemize}
    \item ``model compression'' AND (``knowledge distillation'' OR ``teacher-student'' OR ``feature transfer'')
    \item ``neural network compression'' AND (``knowledge transfer'' OR ``attention transfer'' OR ``parameter reduction'')
    \item ``deep learning'' AND ``model compression'' AND (``integrated gradients'' OR ``explainability'' OR ``interpretability'')
    \item ``efficient neural networks'' AND (``compression factor'' OR ``accuracy-efficiency tradeoff'' OR ``parameter reduction'')
    \item ``knowledge distillation'' AND (``attention map'' OR ``feature representation'' OR ``gradient attribution'')
\end{itemize}

The search was conducted between June 2024 and September 2024. 
Additional relevant papers were identified through citation tracking of key publications. 
Studies were included if they provided quantitative results on compression factors and model accuracy across different datasets. 
We specifically focused on works that presented both teacher and student model architectures with clear parameter counts and performance metrics on benchmark datasets including MNIST, CIFAR-10, CIFAR-100, SVHN, and ImageNet.

The literature review primarily focused on studies published between 2017 and 2024, with particular emphasis on those that achieved high compression factors while maintaining reasonable accuracy, or those that introduced novel distillation techniques relevant to our integrated gradients approach.

\subsubsection{Dataset Characteristics}
Our literature review analysed outcomes across multiple datasets with varying complexity levels. 
MNIST, with its simple structure of 28×28 grayscale handwritten digits across 10 classes, represents the most basic benchmark. 
SVHN (Street View House Numbers) features more complex real-world digit images from street views, introducing challenges like varying illumination, occlusions, and natural backgrounds.
CIFAR-10 and CIFAR-100 represent medium-complexity benchmarks with 32×32 colour images. 
While both share similar image characteristics, the expanded class count of CIFAR-100 (100 vs. 10) introduces additional challenges in inter-class similarity and fine-grained discrimination. 
The inclusion of ImageNet, with its 1000 categories and higher-resolution images, provides a high-complexity benchmark that tests model scalability and feature extraction capabilities across diverse visual concepts.
Several specialised datasets highlight domain-specific applications of knowledge distillation. 
CelebA focuses on facial attributes in celebrity images, testing fine-grained attribute recognition. 
CUHK03 and Market-1501 target person re-identification through surveillance footage, requiring models to match identities across different camera views and conditions. 
DukeMTMCreID similarly addresses re-identification challenges in multi-camera surveillance settings. 
Tiny ImageNet offers a reduced version of ImageNet with 200 classes at 64×64 resolution, serving as an intermediate complexity benchmark. 
Caltech-256, with its 256 object categories in varied poses, scales, and backgrounds, provides another testing ground for object categorisation capabilities.
The varied performance across these datasets illustrates how data complexity influences knowledge distillation outcomes, with simpler datasets generally allowing higher compression factors while maintaining accuracy.

\subsection{Model Architecture and Compression Details}
Our architectural compression strategy preserves critical network components while systematically reducing model size. 
Tables~\ref{tab:compression_details} and \ref{tab:model_blocks} provide complementary views of this process: Table~\ref{tab:compression_details} details the parameter and layer counts across compression levels, outlines the consolidated architecture specifications including attention sources and dividers, and Table~\ref{tab:model_blocks} maps the precise block-level retention strategy across all student models. 
Together, these tables document how we maintained network connectivity and feature extraction capabilities despite substantial parameter reduction.

\subsubsection{Base Model Architecture (Teacher)}

The teacher model (MobileNetV2) was selected for its efficient architecture that balances performance and computational requirements \citep{sandler2018mobilenetv2}. 
MobileNetV2 introduces two key architectural innovations: inverted residual blocks and linear bottlenecks. 
In conventional residual connections, the network uses a wide-narrow-wide pattern, while MobileNetV2 employs a narrow-wide-narrow structure (hence ``inverted'') that expands the input features before applying depth-wise convolution.

The architecture comprises 2,236,682 parameters organised across 53 layers, with 52 convolutional layers followed by a classification head. 
As detailed in Table~\ref{tab:compression_details}, the network processes input through a sequence of modules: an initial ConvBNReLU block, followed by 17 InvertedResidual blocks of varying configurations, and a final ConvBNReLU block before classification. 
Each InvertedResidual block processes data through three sequential steps: (1) an expansion layer that uses point-wise ($1 \times 1$) convolutions to increase the number of channels (e.g., from 24 to 144 channels, a factor of 6 increase), allowing more information to flow in parallel; (2) a depth-wise filtering step that applies $3 \times 3$ convolutions to each channel independently, capturing spatial patterns while keeping computation efficient; and (3) a projection layer that uses point-wise ($1 \times 1$) convolutions to reduce the channels back to a more compact form (e.g., from 144 back to 24 or 32 channels). 
This ``narrow-wide-narrow'' pattern enables the network to process information efficiently while maintaining model performance.

The network gradually increases the feature dimension from the input channels to a 1,280-dimensional feature space before final classification. 
Throughout the network, batch normalisation and ReLU6 activation functions are employed after convolutional layers to improve training stability and introduce non-linearity. 
All operations utilise standard 32-bit floating-point precision (float32) without any reduced-precision optimisations, maintaining compatibility with standard deep learning frameworks while establishing a performance baseline for our compression experiments.

\begin{table}[t]
\centering
\caption{Detailed specifications for each compression level of the student model}
\label{tab:compression_details}
\begin{tabular}{lcccccc}
\hline
Model     & Inverted & Total  & Output   & Compression & Attention & Memory   \\
         & Residual & Layers & Features & Factor      & Source    & Usage \\ \hline
Teacher   & 17       & 53     & 1,280    & 1.00×       & N/A       & 8737 KB (8.53 MB) \\
Student 1 & 15       & 46     & 160      & 2.19×       & 9         & 3982 KB (3.89 MB) \\
Student 2 & 13       & 40     & 96       & 4.12×       & 9         & 2123 KB (2.07 MB) \\
Student 3 & 11       & 34     & 96       & 7.29×       & 9         & 1199 KB (1.17 MB) \\
Student 4 & 9        & 28     & 64       & 12.04×      & 4         & 726 KB (0.71 MB) \\
Student 5 & 7        & 22     & 64       & 28.97×      & 4         & 302 KB (0.29 MB) \\
Student 6 & 5        & 16     & 32       & 54.59×      & 4         & 160 KB (0.16 MB) \\
Student 7 & 3        & 10     & 24       & 139.43×     & 2         & 63 KB (0.06 MB) \\
Student 8 & 1        & 4      & 16       & 1121.71×    & 0         & 8 KB (0.01 MB) \\ \hline
\end{tabular}
\end{table}

\subsubsection{Derivation of the Compressed Student Models}

Our compression strategy systematically reduces the MobileNetV2 architecture by removing InvertedResidual blocks from the end of the network. 
We create eight student models with progressively higher compression factors, grouped into moderate (2.19×-7.29×), high (12.04×-54.59×), and extreme (139.43×-1121.71×) compression levels.

For moderate compression (Student-1 to Student-3), we preserve the core MobileNetV2 structure while removing 2, 4, and 6 InvertedResidual blocks from the end of the network, respectively. 
This approach maintains the network's fundamental feature extraction capabilities while reducing parameter count. 
The attention mechanism for these models derives from the output of InvertedResidual block 9, capturing mid-level feature representations where both low-level patterns and high-level semantics are present.

For high compression (Student-4 to Student-6), we remove 8, 10, and 12 InvertedResidual blocks, requiring a shift in the attention source to earlier layers (block 4) to accommodate the reduced network depth. 
This more aggressive pruning reduces output feature dimensions to 32-64, compared to the original 1,280 in the teacher model.

The extreme compression cases (Student-7 and Student-8) retain only the initial layers of the network. 
Student-7 keeps just 3 InvertedResidual blocks with attention derived from block 2, while Student-8 retains only the first InvertedResidual block with attention from the initial ConvBNReLU layer. 
These models represent compression factors of 139.43× and 1121.71×, respectively, with drastically reduced feature dimensions of 16-24.

Table~\ref{tab:model_blocks} maps the precise block retention strategy across all student models, with attention map production points marked `A'. 
As compression increases, these attention sources necessarily move toward earlier layers to maintain compatible feature dimensions for knowledge transfer.
The attention transfer mechanism ensures that despite their reduced capacity, student models focus on the same critical regions as the teacher when processing inputs.

For each compressed model, we maintain the classification structure with appropriate adjustments to accommodate the reduced feature dimensions. All models end with an adaptive pooling layer followed by a classifier that maps to the 10 CIFAR-10 classes. Table~\ref{tab:compression_details} documents the relationship between compression factor and feature dimensionality, showing how output features decrease from 160 in Student-1 (2.19× compression) to just 16 in Student-8 (1121.71× compression).

\begin{table}[htbp]
\centering
\caption{Overview of block usage across student models. 
A checkmark (\checkmark) indicates that the corresponding block is present in the student model. 
The letter `A' denotes that the attention map is produced at that specific block.
S1-S8 represent the student models with different compression factors (S8 being the highest compression), while T refers to the teacher model. 
This table illustrates how architecture and 
attention production points vary across compression levels.}
\label{tab:model_blocks}
\begin{tabular}{ccccccccccc}
Block ID & Block Type & S8 & S7 & S6 & S5 & S4 & S3 & S2 & S1 & T \\ \hline
0                                                  & ConvBNReLU                                            & A                         & \checkmark & \checkmark & \checkmark & \checkmark & \checkmark & \checkmark & \checkmark & A \\
1                                                  & InvertedResidual                                      & \checkmark & \checkmark & \checkmark & \checkmark & \checkmark & \checkmark & \checkmark & \checkmark & \checkmark\\
2                                                  & InvertedResidual                                      & -                         & A                         & \checkmark & \checkmark & \checkmark & \checkmark & \checkmark & \checkmark & A \\
3                                                  & InvertedResidual                                      & -                         & \checkmark & \checkmark & \checkmark & \checkmark & \checkmark & \checkmark & \checkmark & \checkmark\\
4                                                  & InvertedResidual                                      & -                         & -                         & A                         & A                         & A                         & \checkmark & \checkmark & \checkmark& A \\
5                                                  & InvertedResidual                                      & -                         & -                         & \checkmark & \checkmark & \checkmark & \checkmark & \checkmark & \checkmark & \checkmark\\
6                                                  & InvertedResidual                                      & -                         & -                         & -                         & \checkmark & \checkmark & \checkmark & \checkmark & \checkmark & \checkmark\\
7                                                  & InvertedResidual                                      & -                         & -                         & -                         & \checkmark & \checkmark & \checkmark & \checkmark & \checkmark & \checkmark\\
8                                                  & InvertedResidual                                      & -                         & -                         & -                         & -                         & \checkmark & \checkmark & \checkmark & \checkmark& \checkmark \\
9                                                  & InvertedResidual                                      & -                         & -                         & -                         & -                         & \checkmark & A                         & A                         & A                       & A  \\
10                                                 & InvertedResidual                                      & -                         & -                         & -                         & -                         & -                         & \checkmark & \checkmark & \checkmark & \checkmark\\
11                                                 & InvertedResidual                                      & -                         & -                         & -                         & -                         & -                         & \checkmark & \checkmark & \checkmark & \checkmark \\
12                                                 & InvertedResidual                                      & -                         & -                         & -                         & -                         & -                         & -                         & \checkmark                         & \checkmark & \checkmark\\
13                                                 & InvertedResidual                                      & -                         & -                         & -                         & -                         & -                         & -                         & \checkmark                         & \checkmark & \checkmark\\
14                                                 & InvertedResidual                                      & -                         & -                         & -                         & -                         & -                         & -                         & -                         & \checkmark & \checkmark\\
15                                                 & InvertedResidual                                      & -                         & -                         & -                         & -                         & -                         & -                         & -                         & \checkmark & \checkmark\\
16                                                 & InvertedResidual                                      & -                         & -                         & -                         & -                         & -                         & -                         & -                         & - & \checkmark\\
17                                                 & InvertedResidual                                      & -                         & -                         & -                         & -                         & -                         & -                         & -                         & - & \checkmark\\
18                                                 & ConvBNReLU                                            & -                         & -                         & -                         & -                         & -                         & -                         & -                         &  & \checkmark\\
19                                                 & Classifier                                            & \checkmark & \checkmark & \checkmark & \checkmark & \checkmark & \checkmark & \checkmark & \checkmark & \checkmark\\ \hline
\end{tabular}
\end{table}

\subsection{Implementation Details of the Training Method}
This supplementary material expands on the methods described in the main article---Knowledge Distillation (KD), Attention Transfer (AT), and Integrated Gradients (IG) as data augmentation---providing implementation details, parameter choices used after completing the hyperparameter optimisation, and pseudocode linked to our code at \url{https://github.com/nordlinglab/ModelCompression-IntegratedGradients/PyTorch_CIFAR10/UTILS_TORCH.py}.
We use the same notation as the main article (Equations 1--6) and aim to facilitate reproducibility while explaining our design decisions pedagogically.

The compression strategy is implemented through custom PyTorch model classes. 
The \texttt{SmallerMobileNet} class (Line 243) removes a specified number of layers from the end of the MobileNetV2 feature extractor while preserving the network's core functionality. 
It dynamically identifies the output channels of the last retained convolutional layer by traversing the network architecture in reverse order, then adapts the classifier accordingly.

\subsubsection{Knowledge Distillation Framework}
KD transfers knowledge from a teacher (MobileNetV2) to a student model using softened logits, as defined in Equation (1) of the main article.
The teacher’s precomputed logits $z_t \in \mathbb{R}^{N \times K}$ (where $N$ is batch size, $K = 10$ is the number of CIFAR-10 classes) are stored and accessed via the \texttt{CIFAR10\_KD} dataset class (Line 603), while training is implemented in \texttt{train\_eval\_kd} (Line 392).

\subsubsubsection{Parameter Choices}
\begin{itemize}
\item \textbf{Temperature $T = 2.5 \in [1, 4]$}: Although default is $T = 2.0$ in \texttt{train\_eval\_kd} (Line 398) and \texttt{kd\_loss} (Line 60), our hyperparameter optimisation yielded $T = 2.5$ as optimal.
Based on the hyperparameter optimisation, we chose $T = 2.5$ within the practical range $[1, 4]$ to moderately soften probabilities, balancing the teacher’s confidence with class relationship information (e.g., a logit of 4 becomes $\exp(4/2.5) \approx 4.95$ vs. $\exp(4) \approx 54.6$ at $T = 1$)).
\item \textbf{$\alpha = 0.01 \in [0, 1]$}: While the default is $\alpha = 0.5$ in the code, our optimal value of $\alpha = 0.01$ weights the distillation loss minimally compared to cross-entropy (Line 399), placing stronger emphasis on ground-truth labels than teacher knowledge.
\item \textbf{Learning Rate $lr = 0.001 \in \mathbb{R}^+$}: A standard choice for Adam optimization (Line 397), stable for our student model’s size.
\item \textbf{Epochs = 100 $\in \mathbb{N}$}: Sufficient for convergence on CIFAR-10 (Line 396).
\end{itemize}

\subsubsubsection{Pseudocode}
\begin{algorithmic}[1]
\State \textbf{Input:} Images $x \in [0, 1]^{N \times C \times H \times W}$, labels $y \in \{0, 1, \ldots, 9\}^N$, teacher logits $z_t \in \mathbb{R}^{N \times K}$, student model $S$, $T = 2.5 \in [1, 4]$, $\alpha = 0.01 \in [0, 1]$, epochs = 100 $\in \mathbb{N}$
\For{epoch = 1 to 100}
    \For{each batch $(x, y, z_t)$}
        \State $z_s \in \mathbb{R}^{N \times K} \gets S(x)$ \Comment{Student logits, Line 431}
        \State $L_\mathcal{H} \in \mathbb{R}^+ \gets \text{CrossEntropy}(z_s, y)$ \Comment{Line 438}
        \State $L_\mathcal{KL} \in \mathbb{R}^+ \gets \text{kd\_loss}(z_s, z_t, T)$ \Comment{Line 433, calls Line 60}
        \State $L_{KD} \in \mathbb{R}^+ \gets (1 - \alpha) L_\mathcal{H} + \alpha L_\mathcal{KL}$ \Comment{Eq. (1), Line 439}
        \State Optimise $S$ with $L_{KD}$ \Comment{Line 442}
    \EndFor
\EndFor
\State \textbf{Return:} Trained $S$
\end{algorithmic}

\subsubsubsection{Implementation Notes}
The \texttt{kd\_loss} function (Line 54) computes $L_\mathcal{KL}$ as $\text{KL}(\text{log\_softmax}(z_s/T), \text{softmax}(z_t/T)) \cdot T^2 / N$, where $T^2$ adjusts gradient scale, and $N$ is the batch size.
We precompute teacher logits offline to save computation, assuming MobileNetV2’s 93.9\% accuracy is reliable.

\subsubsection{Attention Transfer}
AT extends KD by aligning student and teacher attention maps, as in Equation (2).
We implement two specialised classes: \texttt{ModifiedTeacher} (Line 327) and \texttt{ModifiedStudent} (Line 347). 
The \texttt{ModifiedTeacher} class divides the network at a point determined by a \texttt{divider} parameter (typically set to 2, 4, 8, or 20 depending on compression level), extracting intermediate feature maps to generate attention maps.

The \texttt{ModifiedStudent} class mirrors this division structure but with fewer layers, controlled by the \texttt{layer} (Line 348) parameter that specifies how many layers to remove from the end. 
Both classes implement a forward pass that returns both logits and attention maps, enabling simultaneous knowledge and attention transfer during training.

\subsubsubsection{Parameter Choices}
\begin{itemize}
\item \textbf{$T = 2.5 \in [1, 4]$, $\alpha = 0.01 \in [0, 1]$}: Consistent with our KD implementation, not the default code values.
\item \textbf{$\gamma = 0.8 \in [0, 1]$}: While the default is $\gamma = 0.5$ (Line 501), our hyperparameter optimisation found $\gamma = 0.8$ to be optimal, placing a relatively high weight on $L_{\text{AT}}$, emphasising the importance of attention map alignment.
\item \textbf{Attention Map Power $p = 2 \in \mathbb{N}$}: In \texttt{attention\_map} (Line 131), $p = 2$ squares activations before averaging, emphasising stronger features (e.g., $0.5^2 = 0.25$, $1^2 = 1$).
\item \textbf{Learning Rate $lr = 0.001 \in \mathbb{R}^+$, Epochs = 100 $\in \mathbb{N}$}: Consistent with KD (Line 497).
\end{itemize}

\subsubsubsection{Pseudocode}
\begin{algorithmic}[1]
\State \textbf{Input:} Images $x \in [0, 1]^{N \times C \times H \times W}$, labels $y \in \{0, 1, \ldots, 9\}^N$, teacher logits $z_t \in \mathbb{R}^{N \times K}$, teacher maps $A_T \in \mathbb{R}^{N \times H' \times W'}$, student $S$, $T = 2.5 \in [1, 4]$, $\alpha = 0.01 \in [0, 1]$, $\gamma = 0.8 \in [0, 1]$
\For{epoch = 1 to 100}
    \For{each batch $(x, y, z_t, A_T)$}
        \State $z_s \in \mathbb{R}^{N \times K}, A_S \in \mathbb{R}^{N \times H' \times W'} \gets S(x)$ \Comment{Line 534}
        \State $L_\mathcal{H} \in \mathbb{R}^+ \gets \text{CrossEntropy}(z_s, y)$ \Comment{Line 544}
        \State $L_\mathcal{KL} \in \mathbb{R}^+ \gets \text{kd\_loss}(z_s, z_t, T)$ \Comment{Line 536, calls Line 60}
        \State $L_{\text{AT}} \in \mathbb{R}^+ \gets \| A_S - A_T \|_2^2$ \Comment{Line 542, calls Line 137}
        \State $L_{Total} \in \mathbb{R}^+ \gets (1 - \alpha) L_\mathcal{H} + \alpha L_\mathcal{KL} + \gamma L_{\text{AT}}$ \Comment{Eq. (2), Line 545}
        \State Optimise $S$ with $L_{Total}$ \Comment{Line 552}
    \EndFor
\EndFor
\State \textbf{Return:} Trained $S$
\end{algorithmic}

\subsubsubsection{Implementation Notes}
Attention maps are $L_2$-normalised (Line 131) to ensure scale invariance.
It is important to note that the teacher model only outputs attention maps from a single layer, which is strategically selected based on the corresponding student model architecture. 
This layer selection ensures that the attention maps from both teacher and student have compatible dimensions for direct comparison. 
As shown in Table~\ref{tab:model_blocks}, the attention map production points (marked with `A') are positioned at corresponding network depths, with attention sources moving toward earlier layers as compression increases. 
The middle layer is chosen as it balances low-level features and high-level semantics in MobileNetV2.
See Figure~\ref{fig:attention_maps_comparison} for examples.

\subsubsection{Integrated Gradients as Data Augmentation}
IG augments data by overlaying attribution maps, as in Equations (3–6).
For computing the Integrated Gradients, we utilised Captum AI's framework to precompute attribution maps, as shown in our \texttt{Compute\_IGs.py} script located at \url{https://github.com/nordlinglab/PyTorch_CIFAR10/Compute_IGs.py}.
We used Captum's \texttt{IntegratedGradients} class with the MobileNetV2 teacher model as the underlying model for computing attributions. 
The overlay functionality is implemented in \texttt{CIFAR10WithIG} (Line 279).

\subsubsubsection{Parameter Choices}
\begin{itemize}
\item \textbf{Baseline $x' = 0 \in [0, 1]^{C \times H \times W}$}: A zero tensor was used in Captum's IG implementation (\texttt{baselines=inputs * 0} in \texttt{Compute\_IGs.py} (Line 69)), representing a black image as a neutral reference point.
\item \textbf{Attribution Aggregation}: After computing the attributions with Captum, we summed the absolute values across the channel dimension (\texttt{torch.sum(torch.abs(attributions), dim=1)} in \texttt{Compute\_IGs.py} (Line 72)) to create a single-channel attribution map highlighting the important regions regardless of direction of influence.
\item \textbf{Scale Factor $s \in [1, 2]$}: Set in \texttt{random\_logarithmic\_scale} (Line 219) with $\text{scale\_min} = 1$, $\text{scale\_max} = 2$, chosen to moderately enhance IG contrast without over-amplification as called in (Line 305).
\item \textbf{Overlay Probability $p = 0.1 \in [0, 1]$}: Although the default is $p = 0.5$ in \texttt{CIFAR10WithIG} (Line 286), our hyperparameter optimisation found $p = 0.1$ optimal, meaning only 10\% of images are augmented with IG overlays during training.
\item \textbf{Blending Weights $0.5, 0.5 \in [0, 1]$}: Equal weighting in Equation (6) (Line 312) preserves $[0, 1]$ range and visibility of IG.
\end{itemize}

\subsubsubsection{Pseudocode}
\begin{algorithmic}[1]
\State \textbf{Input:} Image $x \in [0, 1]^{C \times H \times W}$, precomputed $IG(x) \in \mathbb{R}^{H \times W}$, $p = 0.1 \in [0, 1]$
\State $IG_{\text{scaled}}(x) \in \mathbb{R}^{H \times W} \gets IG(x)^s$, $s \sim \exp(\mathcal{U}[\ln(1), \ln(2)]) \in [1, 2]$ \Comment{Eq. (4), Line 305, calls Line 219}
\State $\hat{IG}(x) \in [0, 1]^{H \times W} \gets \frac{IG_{\text{scaled}}(x) - \min(IG_{\text{scaled}}(x))}{\max(IG_{\text{scaled}}(x)) - \min(IG_{\text{scaled}}(x))}$ \Comment{Eq. (5), Line 306, calls Line 225}
\State $\hat{IG}(x) \in [0, 1]^{C \times H \times W} \gets \text{repeat}(\hat{IG}(x), 3)$ \Comment{Line 307}
\If{rand() < $p$}
\State $x_{\text{augmented}} \in [0, 1]^{C \times H \times W} \gets 0.5 \cdot x + 0.5 \cdot \hat{IG}(x)$ \Comment{Eq. (6), Line 312}
\Else
\State $x_{\text{augmented}} \in [0, 1]^{C \times H \times W} \gets x$ \Comment{Eq. (6), Line 316}
\EndIf
\State \textbf{Return:} $x_{\text{augmented}}$
\end{algorithmic}

\subsubsubsection{Implementation Notes}
For computational efficiency and reliability, Integrated Gradients were calculated using Captum AI's \texttt{IntegratedGradients} class as shown in \texttt{Compute\_IGs.py} (Line 58). 
We used the MobileNetV2 teacher model as the target model for attribution and computed attributions with respect to the true class label of each input. 
Captum internally uses a trapezoidal rule to approximate the path integral in Equation (3).
The precomputed attribution maps were saved as NumPy arrays (\texttt{Captum\_IGs.npy}) and later loaded during the data augmentation process. 
This approach allowed us to avoid recomputing attributions during training, significantly reducing the computational overhead.

\subsection{Attention Map Analysis}
\begin{figure*}
  \centering
  \includegraphics[width=\linewidth]{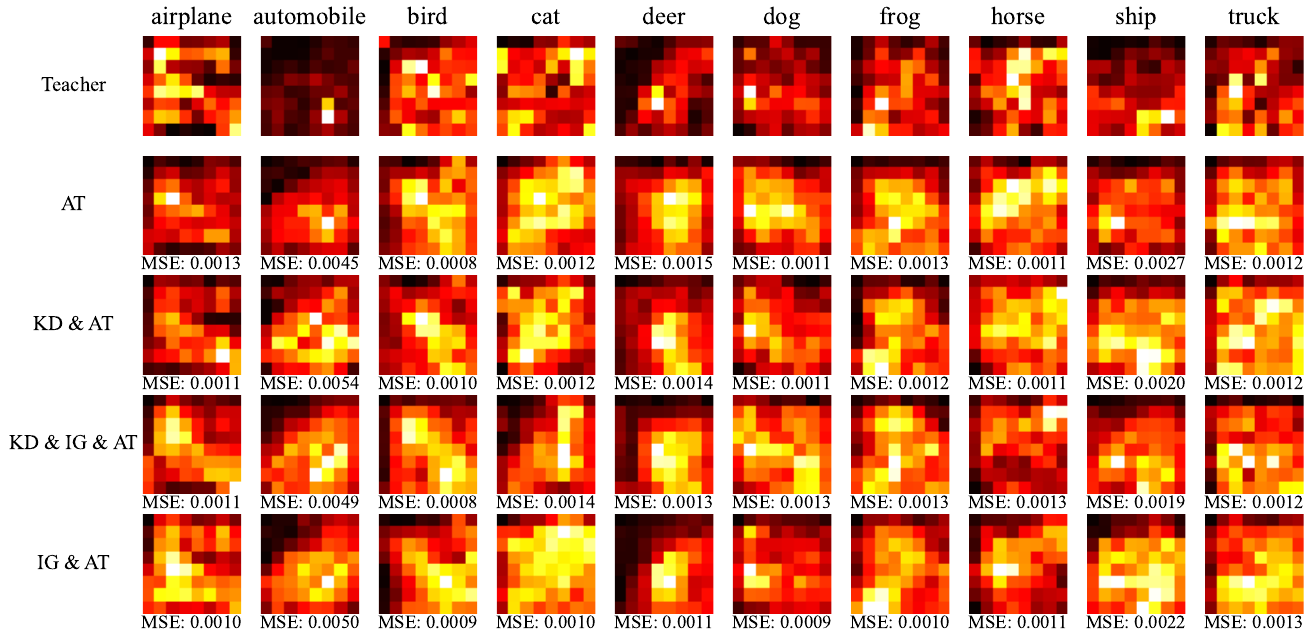}
  \caption{Comparative visualisation of attention maps across different training approaches. Row 1: Teacher model attention maps. Row 2: Student model with attention transfer only (AT). Row 3: Student model using knowledge distillation with attention transfer (KD \& AT). Row 4: Student model using knowledge distillation with integrated gradients and attention transfer (KD \& IG \& AT). Row 5: Student model using integrated gradients with attention transfer (IG \& AT). The MSE values below each image quantify the difference between the student map and the corresponding teacher map. All approaches show visually brighter attention regions compared to the teacher, despite low MSE values, with the automobile class showing the highest discrepancy across methods.}
  \label{fig:attention_maps_comparison}
\end{figure*}

Fig.~\ref{fig:attention_maps_comparison} provides a visual comparison of attention patterns across different model configurations.
The attention maps reveal that while compressed models maintain similar focus patterns to the teacher, they tend to exhibit more concentrated attention regions.
This concentration effect is particularly evident in the automobile class, which consistently shows higher Mean Squared Error (MSE) values compared to other classes.

\subsection{Statistical Analyses}
Our statistical evaluation encompassed several complementary approaches to validate the robustness of our compression method.
 Table~\ref{tab:lilliefors} confirms the normality of our accuracy distributions through Lilliefors tests.

The Monte Carlo simulation results, detailed in Table~\ref{tab:mc_confidence_analysis}, demonstrate how different numbers of simulation runs affect the statistical confidence of our findings.
Additionally, Table~\ref{tab:inference_times} provides comprehensive timing analyses across compression factors, quantifying the computational efficiency gains achieved through our approach.

\begin{table*}[h]
    \centering
    \caption{Lilliefors test results and variance of testing accuracies}
    \label{tab:lilliefors}
    \begin{tabular}{cccc}
        \hline
        Model Configuration & ksstat & p-value & Variance \\ 
        \hline
        Student (baseline) & 0.0862 & 0.3746 & 0.1176 \\
        KD  & 0.0768 & 0.5511 & 0.1134 \\
        KD \& IG & 0.0978 & 0.2024 & 0.3054 \\
        KD \& IG \& AT & 0.0593 & 0.8937 & 0.0881 \\
        \hline
    \end{tabular}
\end{table*}

\begin{table*}[h!]
\centering
\caption{Analysis of Monte Carlo simulation run counts for the KD \& IG model with 4.12 compression factor.
  The table shows how different numbers of runs affect the width of the 95\% confidence interval for accuracy measurements, the relative width compared to our chosen baseline of 60 runs, and the approximate GPU computation time required.}
\label{tab:mc_confidence_analysis}
\begin{tabular}{ccc}
\hline
Number of & 95\% Confidence & Relative Width \\
MC Runs & Interval Width & (60 runs = 100\%) \\
\hline
40 & ±0.19\% & 127\% \\
50 & ±0.17\% & 113\% \\
60 (used) & ±0.15\% & 100\% \\
70 & ±0.14\% & 93\% \\
80 & ±0.13\% & 87\% \\
\hline
\end{tabular}
\end{table*}

\begin{table*}[h]
\centering
\caption{Inference time comparison across different compression factors.
  Times represent the average processing time for a batch of 64 images on our test hardware.}
\label{tab:inference_times}
\begin{tabular}{ccc}
\hline
Compression & Inference Time & Speedup vs. \\
Factor & (seconds/batch) & Teacher Model \\
\hline
1.00× (Teacher) & \num{1.4e-1} & 1.0× \\
2.19× & \num{1.33e-2} & 10.6× \\
4.12× & \num{1.27e-2} & 11.1× \\
7.29× & \num{9.32e-3} & 15.0× \\
12.04× & \num{8.20e-3} & 17.1× \\
28.97× & \num{6.80e-3} & 20.6× \\
54.59× & \num{5.54e-3} & 25.27× \\
139.43x & \num{3.92e-3} & 35.71x \\
1121.71x & \num{1.35e-3} & 103.5x \\
\hline
\end{tabular}
\end{table*}

\subsection{Hyperparameter Optimisation Results}
\begin{figure}[ht]
  \centering
  \includegraphics[width=\textwidth]{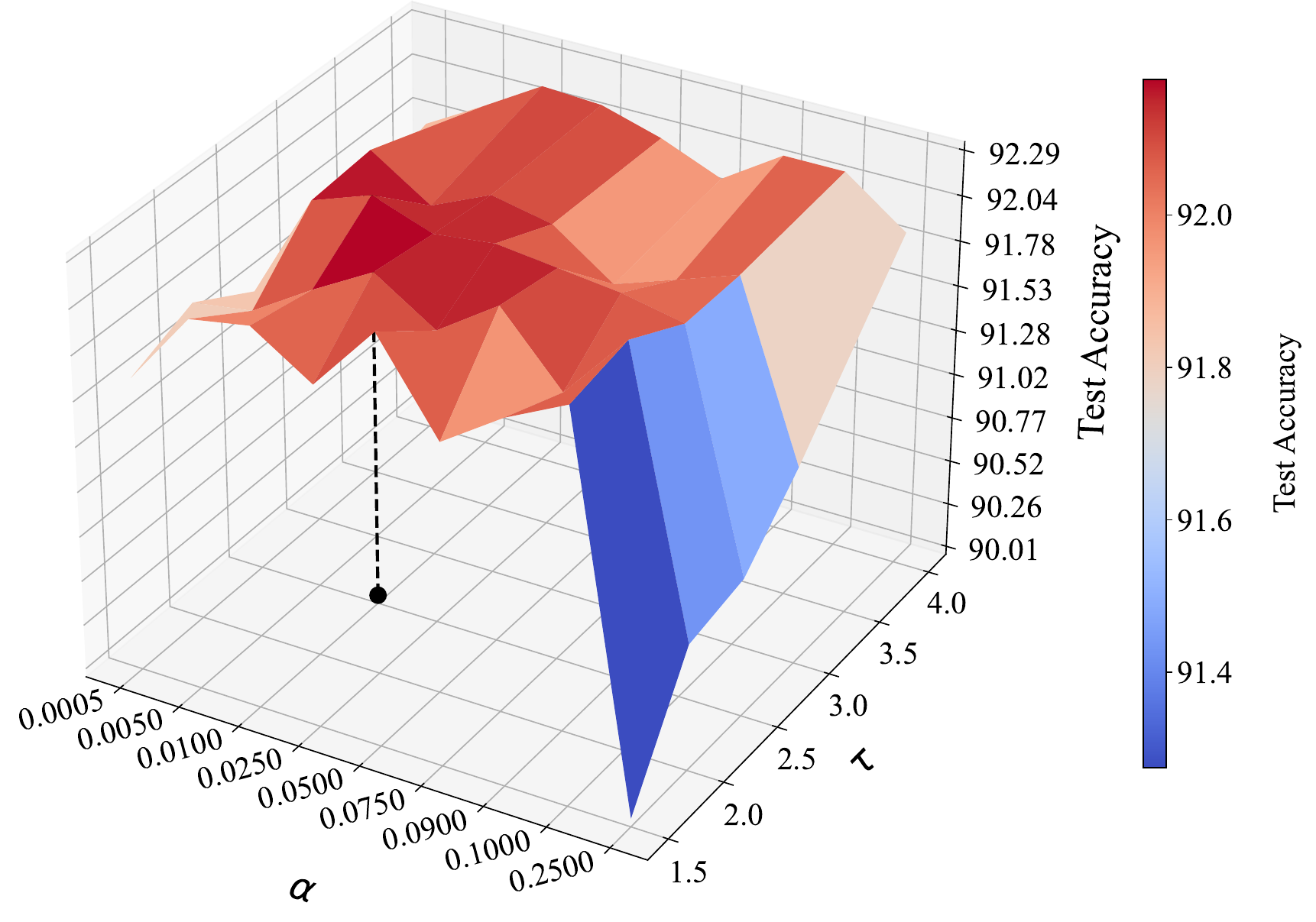}
  \caption{Testing accuracy as a function of weighting parameter \(\alpha\) and temperature \(T\) for the student model with 4.12 compression factor.
  The optimal combination (\(\alpha = 0.01\) and \(T = 2.5\)) achieves 92.29\% accuracy.}
  \label{fig:surface_plot}
\end{figure}

Our hyperparameter search focused on identifying optimal settings for knowledge distillation, integrated gradients, and attention transfer components.
Table~\ref{tab:kd_params} shows the testing accuracies achieved with different combinations of distillation weight parameter $\alpha$ and temperature $T$ for the student model with 4.12 compression factor.
The surface plot in Fig.~\ref{fig:surface_plot} visualises this relationship, highlighting the interaction between these key parameters.

For integrated gradients, Table~\ref{tab:ig_overlay} demonstrates how different overlay probabilities affect model performance.
The results indicate that a probability of 0.1 provides the optimal balance between feature emphasis and model generalisation.
Similarly, Table~\ref{tab:at_params} presents our findings for attention transfer weight optimisation, showing how different $\gamma$ values influence the ability of the model to mimic teacher attention patterns.

\begin{table*}[ht]
\centering
\caption{KD parameter search results showing testing accuracies for various \(\alpha\) and \(T\) combinations for the student model with 4.12 compression factor.}
\begin{tabular}{cccccc}
\hline
\(\alpha\) \textbackslash T & \(1.5\) & \(2\) & \(2.5\) & \(3\) & \(4\) \\
\hline
0.0005 & 91.62 & 91.77 & 91.57 & 91.55 & 91.77 \\
0.005  & 92.04 & 91.82 & 92.17 & 92.20 & 91.96 \\
0.01   & 92.10 & 92.03 & \textbf{92.29} & 91.98 & 92.16 \\
0.025  & 91.88 & 92.22 & 92.17 & 92.13 & 92.14 \\
0.05   & 92.26 & 92.00 & 92.21 & 92.06 & 92.04 \\
\hline
\end{tabular}
\label{tab:kd_params}
\end{table*}

\begin{table*}[ht]
\centering
\caption{IG parameter search results showing the impact of overlay probability on model accuracy for the student model with 4.12 compression factor.}
\begin{tabular}{cc}
\hline
Overlay Probability & Test Accuracy (\%) \\
\hline
0.5  & 91.5 \\
0.25 & 91.99 \\
0.2  & 92.13 \\
0.15 & 92.12 \\
0.1  & \textbf{92.58} \\
0.09 & 92.28 \\
\hline
\end{tabular}
\label{tab:ig_overlay}
\end{table*}

\begin{table*}[ht]
\centering
\caption{Results of the AT parameter search showing the impact of \(\gamma\) on model accuracy for the student model with 4.12 compression factor.}
\begin{tabular}{cc}
\hline
\(\gamma\) & Accuracy (\%) \\
\hline
0.9  & 92.27 \\
0.8  & \textbf{92.42} \\
0.75 & 92.33 \\
0.7  & 92.27 \\
0.6  & 92.20 \\
0.5  & 92.28 \\
\hline
\end{tabular}
\label{tab:at_params}
\end{table*}

\subsection{ImageNet Validation Details}
The evaluation of our models on ImageNet data required careful mapping between CIFAR-10 and ImageNet classes.
Table~\ref{tab:imagenet} provides the complete correspondence between CIFAR-10 categories and their ImageNet counterparts.
This mapping enabled us to assess model generalisation beyond the training domain while maintaining semantic consistency.
To ensure a rigorous evaluation framework, we first assessed the teacher model's performance on our ImageNet subset, analysing its accuracy across all CIFAR-10 class equivalents. 
As shown in Table~\ref{tab:teacher_accuracy}, the teacher model achieved an overall accuracy of 67.37\% (3,537 correctly classified images out of 5,250 total samples). 
However, this performance varied substantially across classes, with notably high accuracy for class 0 (airplane, 84.00\%) and class 3 (cat, 81.20\%), while classes 1 (Automobile) and 5 (Dog) presented significant challenges with accuracies of 53.86\% and 46.27\%, respectively. 
This class-wise variation yielded a balanced accuracy of 68.52\% and Cohen's Kappa of 0.5528. 
To isolate the effects of model compression from inherent classification challenges, we excluded all images incorrectly classified by the teacher model from our subsequent evaluations, resulting in a refined test set of 3,537 images where teacher accuracy is, by definition, 100\%.

\subsubsection{Visualising Integrated Gradients on ImageNet}

\begin{figure}
    \centering
    \includegraphics[width=\textwidth]{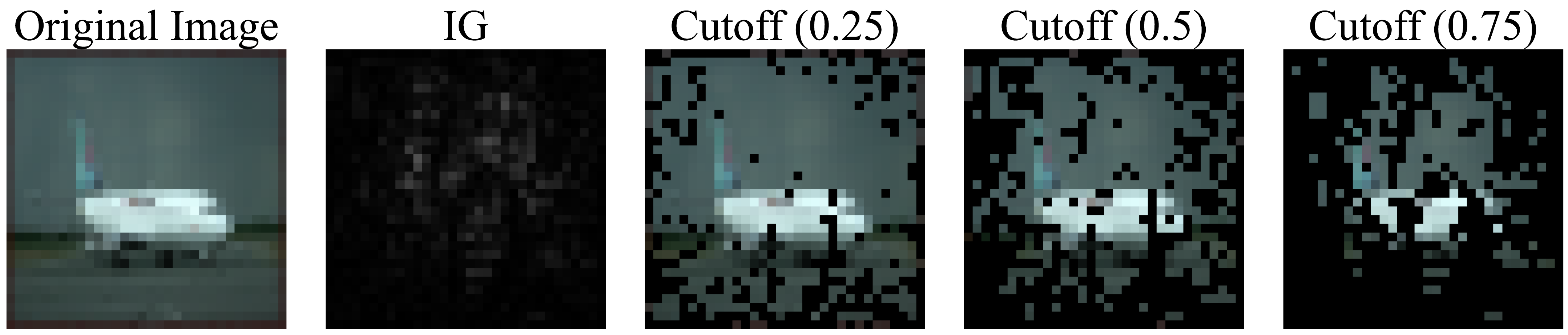}
    \caption{Integrated Gradients visualisation for the teacher model applied to an Airliner image from ImageNet.
    Progressive cutoffs (0.25, 0.5, 0.75) isolate pixels with increasing importance levels, revealing that the model focuses on the distinctive features of the aircraft.}
    \label{fig:imagenet_igs}
\end{figure}

Fig.~\ref{fig:imagenet_igs} provides qualitative insights into how integrated gradients guide the focus of the model when processing complex images.
The visualisations show IG attributions for an Airliner image from ImageNet, with progressive cutoffs applied to isolate the most influential regions.
At the 0.25 cutoff, a broad outline of the aircraft is visible, while the 0.5 and 0.75 cutoffs progressively highlight the most discriminative features, particularly the wings, engines, and fuselage contours.

The pixelated appearance of these maps is a natural consequence of the attribution process and does not indicate a need for smoothing, as the discreteness of the attributions provides precise localisation of feature importance.
The fact that these attributions align well with intuitively important regions for aircraft identification—despite the model being trained on much smaller CIFAR-10 images—demonstrates the robustness of the feature attribution mechanism across image scales and resolutions.

\begin{figure}
    \centering
    \includegraphics[width=\textwidth]{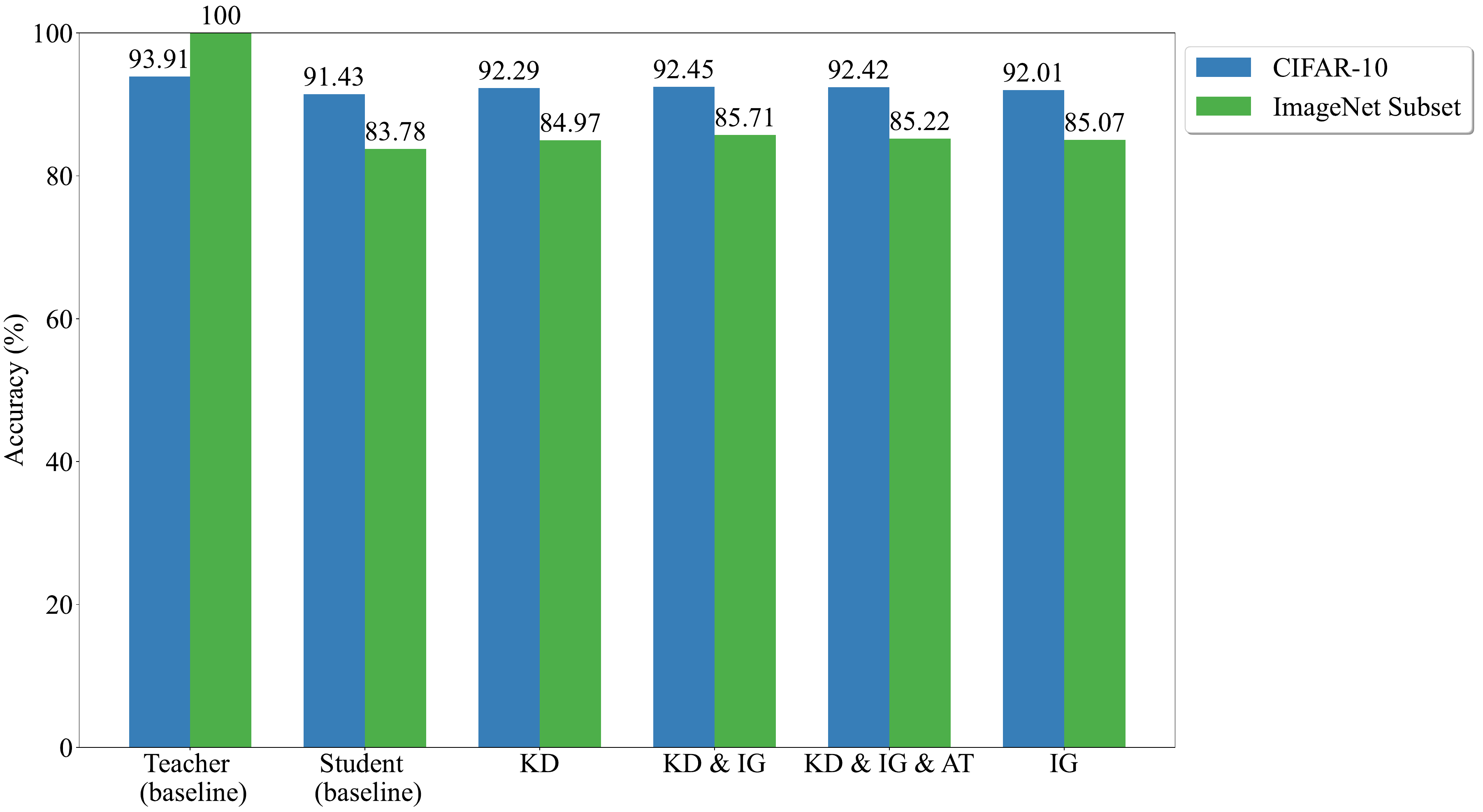}
    \caption{Performance of various model configurations on the ImageNet validation subset for the student model with 4.12 compression factor.
    Both raw accuracy and balanced accuracy (which accounts for class imbalance) are shown.}
    \label{fig:imagenet_bar_chart}
\end{figure}

These visualisations provide a direct window into the decision-making process of the model, confirming that it focuses on class-relevant features rather than spurious correlations or background elements.
This interpretability advantage, combined with the performance benefits demonstrated in our quantitative evaluations, establishes KD \& IG as a powerful approach for creating compressed models that are both accurate and explainable. 
A quantified comparison of the model accuracies on CIFAR-10 and ImageNet is visualised in Fig.~\ref{fig:imagenet_bar_chart}

\begin{table}[h]
\centering
  \caption{Mapping of CIFAR-10 class names to their corresponding ImageNet classes.}
\label{tab:imagenet}
\begin{tabular}{cc}
\hline
\textbf{CIFAR-10 Class} & \textbf{ImageNet Classes} \\
\hline
\multirow{2}{*}{Plane} & Airliner, Warplane, Airship \\
& \\
\multirow{3}{*}{Car} & Sports Car, Passenger Car, Go Kart, Golf Cart, Police Van, \\
& Minivan, Moving Van, Pickup, Model T, Tractor \\
& Cab, Convertible, Limousine, Jeep \\
& \\
\multirow{10}{*}{Bird} & Cock, Hen, Hummingbird, Brambling, Goldfinch, House Finch, \\
& Junco, Indigo Bunting, Robin, Bulbul \\
& Jay, Magpie, Chickadee, Water Ouzel, Kite, Bald Eagle, Vulture, \\
& Great Grey Owl, Black Grouse, Ptarmigan \\
& Ruffed Grouse, Prairie Chicken, Peacock, Quail, Partridge, African Grey, \\
& Macaw, Sulphur-Crested Cockatoo, Lorikeet, Coucal \\
& Bee Eater, Hornbill, Jacamar, Toucan, Drake, Red-Breasted Merganser, \\
& Goose, Black Swan, White Stork, Black Stork \\
& Spoonbill, Flamingo, American Egret, Little Blue Heron, Bittern, \\
& Crane, Limpkin, American Coot, Bustard, Ruddy Turnstone \\
& Red-Backed Sandpiper, Redshank, Dowitcher, Oystercatcher, \\
& European Gallinule, Pelican, King Penguin, Albatross \\
& \\
\multirow{1}{*}{Cat} & Egyptian Cat, Persian Cat, Tiger Cat, Tabby Cat, Siamese Cat \\
& \\
\multirow{1}{*}{Deer} & None \\
& \\
\multirow{3}{*}{Dog} & Siberian Husky, Malamute, Ibizan Hound, Whippet, Beagle, \\
& Labrador Retriever, Curly-Coated Retriever, Giant Schnauzer, \\
& Standard Schnauzer, Lhasa, West Highland White Terrier, Cairn, \\
& German Shepherd, Golden Retriever, Lakeland Terrier \\
& \\
\multirow{1}{*}{Frog} & Tailed Frog, Bullfrog, Tree Frog \\
& \\
\multirow{1}{*}{Horse} & None \\
& \\
\multirow{1}{*}{Ship} & Yacht, Aircraft Carrier, Schooner, Lifeboat, Canoe \\
& \\
\multirow{1}{*}{Truck} & Trailer Truck, Garbage Truck \\
& \\
\hline
\end{tabular}
\end{table}

\begin{table}[htbp]
\centering
\caption{Teacher model performance on ImageNet subset by class, showing the number of correctly classified samples and exclusions}
\label{tab:teacher_accuracy}
\begin{tabular}{lccc}
\hline
Class & Accuracy (\%) & Correct Samples & Total Samples \\
\hline
Airplane (0) & 84.00 & 126 & 150 \\
Automobile (1) & 53.86 & 377 & 700 \\
Bird (2) & 73.28 & 2,125 & 2,900 \\
Cat (3) & 81.20 & 203 & 250 \\
Deer (4) & 0.00 & 0 & 0 \\
Dog (5) & 46.27 & 347 & 750 \\
Frog (6) & 66.00 & 99 & 150 \\
Horse (7) & 0.00 & 0 & 0 \\
Ship (8) & 77.60 & 194 & 250 \\
Truck (9) & 66.00 & 66 & 100 \\
Total & \textbf{67.37} & \textbf{3,537} & \textbf{5,250} \\
\hline
\end{tabular}
\end{table}

\subsection{Detailed Performance Metrics Across Compression Levels}
Table~\ref{tab:compression_performance} provides comprehensive performance metrics across different compression factors and training configurations. 
These results demonstrate how various components of our approach contribute to final model performance and how compression affects different aspects of model behaviour.
Table~\ref{tab:compression_performance} presents the detailed testing accuracies for each configuration, enabling precise comparison of different approaches across compression levels.

\subsubsection{Computational Performance Analysis}

The computational efficiency of our approach extends beyond model size reduction. 
Fig.~\ref{fig:inference_vs_acc} provides comprehensive timing analyses across compression factors, quantifying the computational efficiency gains achieved through our approach. 
The 4.12× compressed student model reduces inference time from 0.140s to 0.013s per image batch, representing a 10.8× speedup that exceeds the compression ratio. 
This non-linear improvement in computational efficiency likely results from reduced memory access patterns and better cache utilisation in the smaller model.
Our precomputation strategy for integrated gradients proves particularly effective, reducing the training overhead compared to computing IG during each iteration. 
While the initial precomputation requires approximately 2 hours for the CIFAR-10 training set, this one-time cost eliminates what would have been an additional 40+ hours of computation during the student training process. 
This trade-off becomes even more favourable for scenarios requiring multiple training runs or hyperparameter searches.

\begin{figure}[h]
\centering
\includegraphics[width=\textwidth]{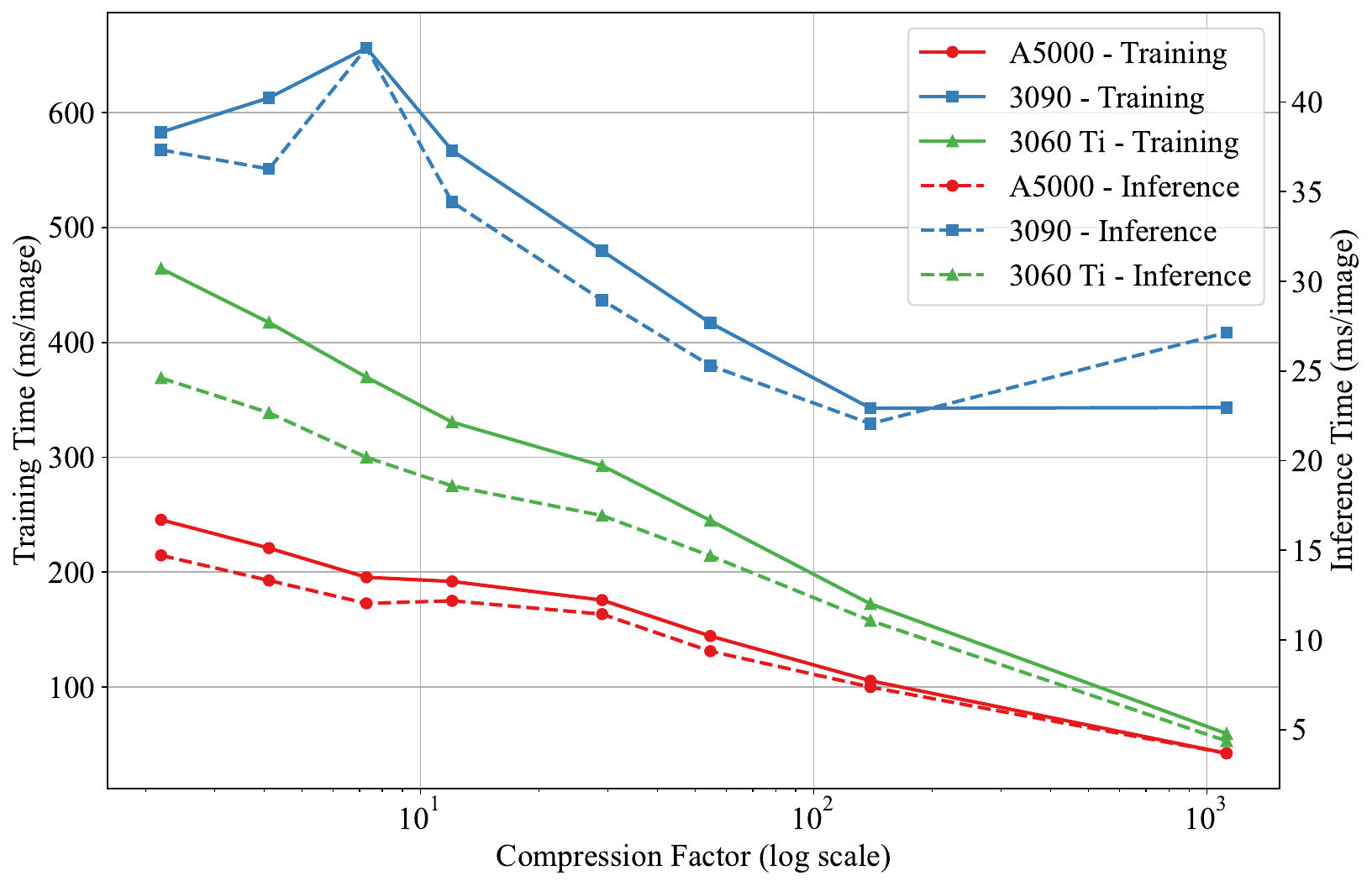}
\caption{
Training and inference times as a function of compression factor across three different GPU configurations. 
The left y-axis shows training time (ms/image) represented by solid lines, while the right y-axis shows inference time (ms/image) represented by dashed lines. 
Both axes use linear scale while the x-axis (compression factor) uses logarithmic scale. 
The results demonstrate that higher compression factors lead to significant reductions in both training and inference times across all hardware configurations, with the A5000 showing the best performance overall.
}
\label{fig:gpu_performance}
\end{figure}

Figure~\ref{fig:gpu_performance} provides a comprehensive analysis of both training and inference times across different compression factors on three GPU hardware configurations (NVIDIA RTX A5000, RTX 3090, and RTX 3060 Ti). 
As the compression factor increases, both training and inference times decrease significantly, with the highest compression factor (1121.71×) showing approximately a 10× reduction in training time and over 100× reduction in inference time compared to the teacher model. 

Notably, the relationship between compression factor and computational efficiency is non-linear, with some anomalies visible in the performance curves. 
These deviations from the expected linear scaling can be attributed to several factors. 
First, the shared computing infrastructure used for these experiments meant that GPUs were occasionally utilised by other researchers concurrently, introducing variability in the measurements. 
This is particularly evident in the RTX 3090 performance curve, which shows unexpected fluctuations at compression factors between 12.04× and 139.43×.

The non-linear behaviour observed in the RTX 3090 configuration likely stems from data transfer bottlenecks between CPU and GPU memory. 
As model size decreases with higher compression, the limiting factor shifts from GPU computation to memory transfer operations, where CPU speed and system RAM bandwidth become increasingly influential. 
This hypothesis is supported by the convergence of performance across all three hardware configurations at the highest compression levels, suggesting that at extreme compression factors, the models become so small that even the less powerful RTX 3060 Ti can process them with similar efficiency to the high-end A5000.

The A5000 consistently demonstrates the best performance across all compression factors, while the relative performance gaps between different GPU models narrow at higher compression levels, suggesting that highly compressed models can run efficiently even on less powerful hardware. 
This democratising effect of model compression highlights its potential for deploying sophisticated models on resource-constrained devices.

\begin{figure}
    \centering
    \includegraphics[width=\textwidth]{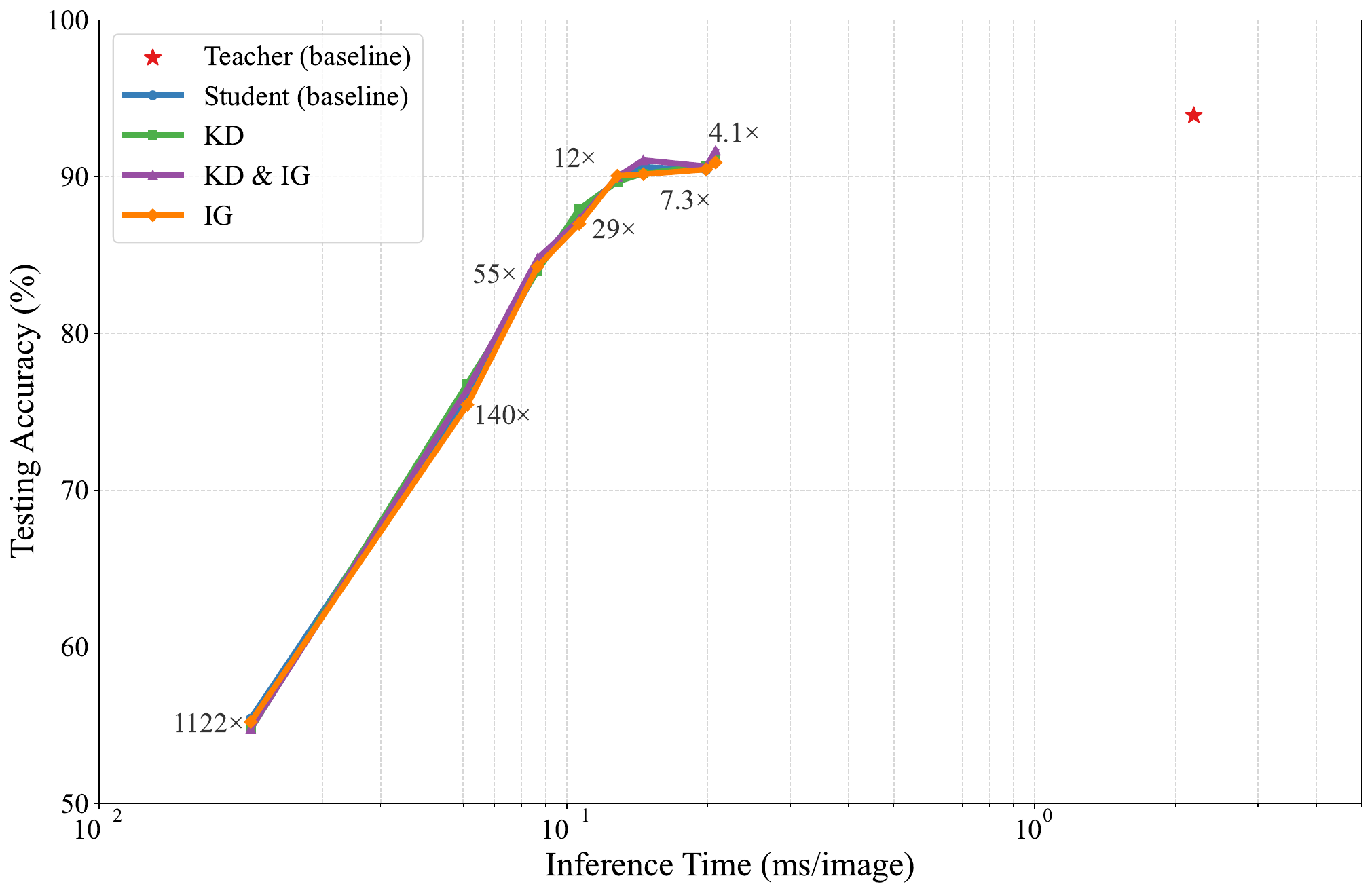}
    \caption{Performance of various model configurations on the ImageNet validation subset for the student model with 4.1 compression factor.
    Both raw accuracy and balanced accuracy (which accounts for class imbalance) are shown.}
    \label{fig:inference_vs_acc}
\end{figure}

\begin{table*}[ht!]
\centering
\caption{Highest testing accuracies (\%) across different compression factors and training configurations. Best results for each compression factor are highlighted in bold.}
\label{tab:compression_performance}
\begin{tabular}{ccccc}
\hline
Compression & \multicolumn{4}{c}{Testing Accuracy (\%)} \\
Factor & Student & KD & IG & KD \& IG \\
\hline
2.19x & 91.66 & 92.05 & 92.08 & \textbf{92.18} \\
4.12x & 91.50 & 92.29 & 92.01 & \textbf{92.58} \\
7.29x & 91.35 & 91.68 & 91.51 & \textbf{92.00} \\
12.04x & 90.19 & 90.73 & 90.90 & \textbf{90.94} \\
28.97x & 87.97 & 88.02 & 87.84 & \textbf{88.16} \\
54.59x & 84.91 & 85.13 & \textbf{85.20} & 84.84 \\
139.43x & 78.61 & \textbf{79.13} & 78.47 & 78.24 \\
1121.71x & 55.48 & 55.64 & \textbf{55.82} & 55.03 \\
\hline
\end{tabular}
\end{table*}

\newpage
 \bibliography{References}